\documentclass[letterpaper]{article} 
\usepackage{aaai24}  
\usepackage{times}  
\usepackage{helvet}  
\usepackage{courier}  
\usepackage[hyphens]{url}  
\usepackage{graphicx} 
\usepackage{natbib}  
\usepackage{caption} 
\DeclareUnicodeCharacter{2212}{\ensuremath{-}}
\usepackage{algorithm}
\usepackage{algorithmic}

\usepackage{newfloat}
\usepackage{listings}
\DeclareCaptionStyle{ruled}{labelfont=normalfont,labelsep=colon,strut=off} 
\floatstyle{ruled}
\newfloat{listing}{tb}{lst}{}
\floatname{listing}{Listing}

\usepackage{booktabs,caption}

\usepackage[flushleft]{threeparttable}

\usepackage{amsmath}

\usepackage{relsize}
\usepackage{amssymb}
\usepackage{amsfonts}
\newtheorem{definition}{Definition}

\newtheorem{remark}{Remark}

\usepackage{color, colortbl}
\definecolor{Gray}{gray}{0.9}

\usepackage[utf8]{inputenc}

\DeclareMathAlphabet\mathcal{OMS}{cmsy}{b}{n}
\DeclareMathAlphabet\mathbfcal{OMS}{cmsy}{b}{n}

\title{Curriculum Learning and Imitation Learning for Model-free Control on Financial Time-series}
\author{
    Woosung Koh\textsuperscript{\rm 1}\equalcontrib,
    Insu Choi \textsuperscript{\rm 2}\equalcontrib,
    Yuntae Jang\textsuperscript{\rm 1}\equalcontrib,
    Gimin Kang\textsuperscript{\rm 2},
    Woo Chang Kim\textsuperscript{\rm 2}
}
\affiliations {
    \textsuperscript{\rm 1} Yonsei University\\
    \textsuperscript{\rm 2} Korea Advanced Institute of Science and Technology\\
    \{reiss.koh, jytfdsa1218\}@yonsei.ac.kr, \{jl.cheivly, kgm4752, wkim\}@kaist.ac.kr
}

\begin{document}

\maketitle

\begin{abstract}
Curriculum learning and imitation learning have been leveraged extensively in the robotics domain. However, minimal research has been done on leveraging these ideas on control tasks over highly stochastic time-series data. Here, we theoretically and empirically explore these approaches in a representative control task over complex time-series data. We implement the fundamental ideas of curriculum learning via data augmentation, while imitation learning is implemented via policy distillation from an oracle. Our findings reveal that curriculum learning should be considered a novel direction in improving control-task performance over complex time-series. Our ample random-seed out-sample empirics and ablation studies are highly encouraging for curriculum learning for time-series control. These findings are especially encouraging as we tune all overlapping hyperparameters on the baseline---giving an advantage to the baseline. On the other hand, we find that imitation learning should be used with caution.
\end{abstract}

\section{Introduction}

By the end of 2020, the total assets under management reached $\$100$ trillion U.S. dollar mark \citep{heredia2021100}. Optimizing investment portfolios and trading the markets has been an ongoing challenge, especially as literature has criticized human managers' discretionary management of funds \citep{fama1995random}. With the immense capital at risk and heightened competition for management fees, optimizing investment decisions continues to be an active area of research for the machine learning and control discipline \citep{gupta2019hybrid, ma2021portfolio, pinelis2022machine}.

Despite the interest, a fundamental challenge plagues the financial control domain---fixed access to the data-generating process $p_{data}$. Concretely, the training data $\hat{p}_{data}$, which is a sampled approximation to $p_{data}$, is fixed and can not be further sampled without the passage of time.

We draw parallels to the robotics domain. The $p_{data}$ in physical systems encompass the manipulator and the manipulated object(s). The sample size of $\hat{p}_{data}$ can be raised arbitrarily via real simulations \citep{kalashnikov2018scalable, kalashnikov2021scaling}, or physics simulators running in parallel \citep{9196808, makoviychuk2021isaac}.

This is unavailable to the financial control domain as we must work with a unique set of input and output features of interest, and its mapping distribution we approximate can only be sampled over the temporal dimension. Each joint stochastic processes of financial variables are distributed uniquely. $\hat{p}_{data}$ of interest can not be reasonably proxied by other variables where more samples of $\hat{p}_{data}$ may exist.

This results in a domain-specific need for improved signal learning and optimization with limited noisy data. Here, we explore two highly leveraged approaches in the physical control domain: (i) Curriculum Learning (CL) and (ii) Imitation Learning (IL) to best use the fixed samples of $\hat{p}_{data}$. Despite the integral nature of CL and IL in the physical control domain \citep{kilinc2021follow, Berg-RSS-23, Haldar-RSS-23, Reuss-RSS-23}, it has been rarely explored for control over financial time-series.

To leverage these paradigms, we work with model-free Reinforcement Learning (RL) \citep{degris2012model}---ensuring that our approach universally applies to all financial control tasks. By bridging the gap between the ample IL and CL approaches in the physical control domain and the financial control domain, we discover noteworthy improvements in generalized performance.

Through the theoretical and empirical study, our contributions are as follows:

\begin{itemize}

\item To the best of our knowledge, the first curriculum learning approach for financial control
\item Extensive out-sample empirical results that beat up-to-date baselines for all three underlying algorithms
\item Demonstrated on two representative data sets simulating inter-asset-class and intra-asset-class problems, with each optimization problem constrained differently to simulate real-world problems
\item The statistical significance of our approach is especially notable as the overlapping hyperparameters are tuned for the baseline---giving an advantage to the baselines
\item Analysis of curriculum learning and imitation learning approaches under a unified statistical framework of signal and noise decomposition

\end{itemize}

\section{Related Works}

\subsection{Curriculum Learning}

Curriculum Learning (CL) has been a powerful tool in training deep learning and RL systems. Inspired by teaching a human student, the data set is augmented s.t. the network is first exposed to easier data and sequentially exposed to more challenging data \citep{wang2021survey}. This framework has been applied to training non-sequential tasks, such as computer vision classification \citep{hacohen2019power}, and more commonly used on sequential tasks that leverage RL \citep{portelas2021automatic, narvekar2020curriculum}. Recently, a work has leveraged CL in forecasting financial revenue \citep{koenecke2020curriculum}. To the best of our knowledge, no literature explores the potential for CL in high noise-to-signal complex financial systems---i.e., the financial market. Specifically, it does not leverage CL in an end-to-end control-task problem.

\subsection{Imitation Learning}

Imitation Learning (IL), synonymous with Learning from Demonstrations (LfD) and Behaviour Cloning (BC), is a critical approach in the robotics control learning domain \citep{hsiung2022learning}. In short, IL and its variants aim to converge to an optima for the \textit{student} (an agent we want to train) via imitating the behavior of a \textit{teacher}, also called \textit{expert}, or \textit{oracle} \citep{hsiung2022learning}. The extensive literature on IL in autonomous physical control highlights the strengths of IL \citep{ravichandar2020recent}. Concretely, IL is highly useful when there is no tractable label generator with enough fidelity for training. IF is also highly sample efficient as it allows the student to quickly converge its learned parameters to a near-optimal region in the high-dimensional parameter space. Unlike CL, more literature on leveraging IL does exist in financial control. Representative papers include \citep{liu2020adaptive} and \citep{fang2021universal}. Liu et al.'s work imitates two experts: (i) a heuristic expert and (ii) an oracle expert with access to future data. On the other hand, the work of Fang et al. only imitates an oracle. However, as mentioned in \citep{goluvza2023imitation}, there lacks depth in exploring IL as a feasible approach for financial control.

\subsection{RL for Financial Control}
RL has made significant strides in financial control since 2020, leveraging large data sets and neural-networks to improve financial decision-making without relying on model assumptions. \citep{hambly2023recent} outlines RL's applications in various financial tasks and its integration with deep learning techniques, highlighting the method's growing relevance in navigating the complexities of financial markets. \citep{charpentier2021reinforcement} presents up-to-date overview of RL techniques with applications in economics and finance. They illustrate how RL, combined with computational advances in deep learning, can address complex behavioral problems in these fields.

\section{Preliminary: Signal and Noise}
All sequential control tasks vis-à-vis interacting with public financial markets suffer from high stochasticity and the accompanying high noise-to-signal ratio. This can be caused by (i) innate stochasticity in the system being modeled, (ii) incomplete observability, and (iii) incomplete modeling. Modeling the financial market suffers from all three fundamental drivers. Notably, it is evident that incomplete observability and modeling are major causes. Financial markets absorb external shocks, i.e., real-world events, via information flow across time and variables. Therefore, perfectly modeling this data-generating process $p_{data}$ requires a world model. Additionally, we can not include every financial variable in our model due to the intractably high-dimensional feature count. Even if this was feasible, there is little guarantee that it would be helpful for our model---which is very likely modeling a finite set of variables of interest. This signal and noise theory fundamentally drives our entire approach.

Denote a movement in a tradable security $sec$, $\Delta sec_t = sec_t - sec_{t-1} \approx \ln({sec_t/sec_{t-1}})$. Here we work with a discrete time space denoted $t \in \mathbfcal{T} := \{1, ..., \vert \mathbfcal{T} \vert \}$. Theoretically, we can decompose this movement into signal and noise, $\Delta sec = \Delta signal + \Delta noise$. Here, the signal is defined as the movement in $sec$ that can be reliably approximated via mapping the given set of signals within our feature space (which is naturally observable) to future movements less $\Delta noise$. That is, given a noise-free state $ \mathbfcal{S}_{T'}^{nf} := \{ \mathbfcal{SIG}_{t-1},...,\mathbfcal{SIG}_{t-T'}\} $, where $T' \in \mathbb{N}$, $\mathbfcal{SIG}_t := \{ \Delta signal^1_t, ..., \Delta signal^M_t \}$, $m \in \mathbb{N}$, $ \exists 
 \mathcal{F} \ s.t. \ \mathcal{F}: \mathbfcal{S}_{T'}^{nf} \longmapsto \{\Delta signal_t^1,..., \Delta signal_t^M\}  $, where an expressive $\mathcal{F}$ approximator, $\hat{\mathcal{F}}$ performs reasonably well. Naturally, $\Delta noise_t$ is i.i.d. distributionally symmetrical system noise. The theoretical existence of a reasonable mapping function is salient as it allows us to theoretically decompose market movements $\Delta \mathbfcal{M} := \bigcup_{ sec \in \mathbfcal{M} } \{\Delta sec\}$ into movements that influence future time steps, influenced by previous time steps, and the remaining movements that are noise.

Fundamentally, as noise-to-signal $\triangleq \Delta noise/ \Delta signal$ rises, it is more challenging to train gradient-based learning models—i.e., sample efficiency and accuracy fall while variance rises. The challenge with modeling stochastic time series data is that $\exists$ noise in the input $(t-n)$ and label $(t)$ space. Eliminating the noise is challenging as signal and noise decomposition can only be approximated as a posteriori. That is, accurately decomposing $(t-n)$ data before $(t)$ is available is not possible as we can not evaluate $\hat{\mathcal{F}}$. Our methods examine possible ways of reducing the $\Delta noise$ in training the control model while minimally impacting the $\Delta signal$.

\section{Method}
\subsection{Portfolio Control as a Markov Decision Process}

A granular range of sequential financial control tasks exists. These tasks fall under the umbrella of trading, portfolio optimization, optimal execution, or a combination of these. All these tasks interact with the market via increasing, decreasing, or maintaining a position on a discrete universe set, $\mathbfcal{U} := \{sec_1, ..., sec_n, ..., sec_N\}$ where $n \in \mathbb{N}$, via a discrete action set, $\mathbfcal{A} := \{a_1, ..., a_M \ \vert \ a_m \in \{1:long, \ 0:none, \ -1:short\}, m \in \mathbb{N} \}$ and discrete quantity set, $\mathbfcal{Q} := \{q_1, ..., q_N \ \vert \ q \ge 0\}$. Let $\mathbfcal{P} := \{price_{1}, ..., price_N \ \vert \ price_n \ corresponds \ to \ sec_n \in \mathbfcal{U}\}$.
When sets $\mathbfcal{Q}$, $\mathbfcal{A}$, and $\mathbfcal{P}$ are expressed as matrices with appropriate dimension space, $\forall t \in \mathbfcal{T} \setminus \vert \mathbfcal{T} \vert $, $\exists$ transition model $\mathbb{T}: \mathbfcal{Q}^T\mathbfcal{A}^T \ \mathbfcal{P} \longmapsto \mathbfcal{S}_{t+1}$. In the case of portfolio optimization, it is assumed that the portfolio size and in turn, the trading volume is sufficiently large relative to $p_n \in \mathbfcal{P}$ s.t. the optimization problem can be represented via continuous action space, simplifying the problem into $\mathbb{T}: \mathbfcal{W}^T\mathbfcal{P} \longmapsto \mathbfcal{S}_{t+1}$, where typically the weight set, $\mathbfcal{W} = \{w_1, ..., w_N \vert \sum_{N} w_n = 1\}$. Deviations in $\mathbfcal{W}$ depend on the availability of shorting and acceptable financial leverage defined by the modeler. Concretely, for a sufficiently large volume,  $\vert \mathbfcal{Q}^T\mathbfcal{A
} \vert \gg 0$, such as $\vert \mathbfcal{Q}^T\mathbfcal{A} \vert \longrightarrow \infty $, the discrete interaction with the market can be closely approximated via a continuous interaction model. A key advantage to modeling trading with weights is the ease of incorporating portfolio constraints. It is often the case that a modeler's capital is finite and limited. Integrating these limits is convenient in weight-based optimization as it is implicitly modeled. We take advantage of this approach. 

Ideally, the Markov Decision Process (MDP)’s state, $\mathbfcal{S}^{MDP}$, includes all relevant variables that contribute as a signal for optimal policy, $\pi^*$. However, with the virtually infinite variable combinations that could potentially make up $\mathbfcal{S}^{MDP}$ the optimal state space, $\mathbfcal{S}^{MDP*}$ is intractable. Instead, discretionarily including relevant variables based on domain knowledge is common. Commonly, elements of $\mathbfcal{S}^{MDP}$ include lagged information on $\mathbfcal{U}$ and other relevant variables. The MDP details are as in Table 1, unless stated otherwise. The Macro ETFs environment's $\mathbfcal{S}$ includes lagged values $\mathbfcal{U}_{T'}$ and major macroeconomic financial variables $\mathbfcal{M}_{T'}$. The exact state representation fed into the neural-network is available in Appendix Section 2.

\begin{table*}
\centering
\begin{threeparttable}
  \caption{Markov decision process}
  \label{tab:commands}
  \begin{tabular}{|c|cc}
        \toprule
    Environment & Macro ETFs & Commodity Futures \\
    \midrule
       Universe, $\mathbfcal{U}$ & $\{ ETF_1, ..., ETF_4 \}$& $\{ F_1, ..., F_8 \}$ \\
        State, $\mathbfcal{S}$ & $\{ \mathbfcal{U}_{T'}, \mathbfcal{M}_{T'} \}$ & $\{ \mathbfcal{U}_{T'}\}$\\
        Action, $\mathbfcal{A}$ & $\{ a_1, ..., a_4 \ \vert \ a \in \{-1, 0, 1\}\}$ & $\{ a_1, ..., a_8 \ \vert \ a \in \{-1, 0, 1\} \}$  \\
    Reward, $r \in\mathbfcal{R}$ & $\ln({portfolio_t/portfolio_{t-1}})$ & $\ln({portfolio_t/portfolio_{t-1}})$\\
    \bottomrule
  \end{tabular}
    \end{threeparttable}
\end{table*}

\subsection{Imitation Learning for Financial Control}

In a single-learner setting, we call vanilla learner—it must map $ \pi_\theta^{vanilla}: \mathbfcal{S}_{T'} \longmapsto \mathbfcal{A}, \ s.t. \ \mathop{arg 
\ max}_{\theta} \mathbb{E}( r \in \mathbfcal{R} \vert \mathbfcal{S}_{T'}, \pi_\theta )$. This mapping can be decomposed into two stages: (i) temporal signal representation and learning and (ii) output vector optimization given portfolio constraints. The first stage represents the implicit learning of the future state—i.e., learning the causal signal. The second stage represents forming the output vector $\mathbfcal{A}$ s.t. it optimizes conforming to the environment setting---primarily, the portfolio constraint. 

E.g., in the first, upstream stage, a na\"ive $\pi$, $\pi ^ n$ notices that given a certain state vector direction and magnitude, in the next time step $sec_1$, $sec_2$, $sec_3$, and $sec_4$ rise. To a $\pi ^ n$ this signals a long (buy) position allocation $\forall sec$  resulting in a na\"ive signal: $\{1, 1, 1, 1\}$ corresponding to $w(a) := 1/4, \ \forall a \in \mathbfcal{A}$, weight allocation $ \forall sec$, given that a linear normalization is applied to conform to portfolio constraints. A more sophisticated $\pi$, will incorporate the second, downstream stage. Within the neural-network it will transform this signal to a vector $\mathbfcal{A}$ s.t. it maximizes the reward while conforming to the constraints of the environment. Concretely, a sophisticated $\pi$ will learn that too many buy signals dilute the weight, and in turn choose to output a signal that is more concentrated for higher return probability $sec \ s.t. \ \mathop{arg \ max}_{ \theta} \mathbb{E}( r \in \mathbfcal{R} \vert \mathbfcal{S}_{T'} , \pi_\theta) $. E.g., given that the signal suggests higher confidence of $sec_1$ rising, it may $\mathbfcal{A} := \{1, 0, 0, 0\}$.

In the imitation learning paradigm, these two components are detached into two different learners. The oracle $\phi$, first identifies the $\pi^\phi = \pi^*$, which conforms to environmental constraints. $\phi$ can trivially identify $\pi^*$ as it has access to future data. The student learner incorporating $ a^* \in \pi ^* $ can directly learn to conform to an environment-constrained optimal policy. Simply put, the teacher strictly learns stage (ii) while the student focuses on stage (i). By delegating the two stages to two distinct learners, the student learner, in theory, can more efficiently tune its high-dimensional parameters $\theta$. The intuition is visualized in Figures 1 and 2.

\begin{figure}
    \centering
    \includegraphics[width=1\linewidth]{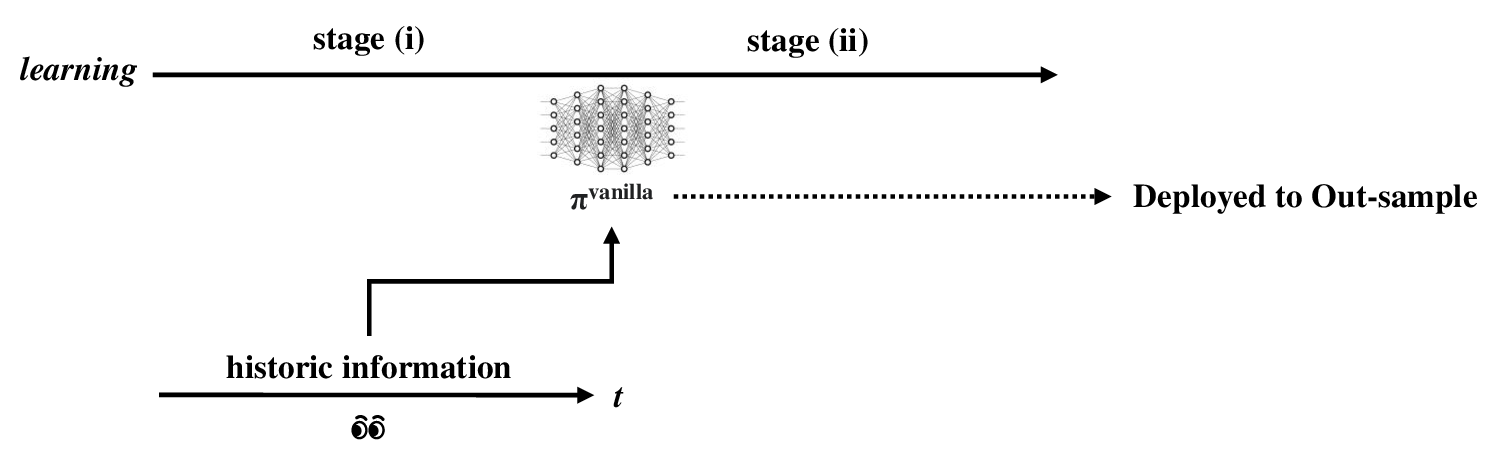}
    \caption{Training an end-to-end vanilla learner}
    \label{fig:Training an end-to-end vanilla learner}
\end{figure}
\begin{figure}
    \centering
    \includegraphics[width=1\linewidth]{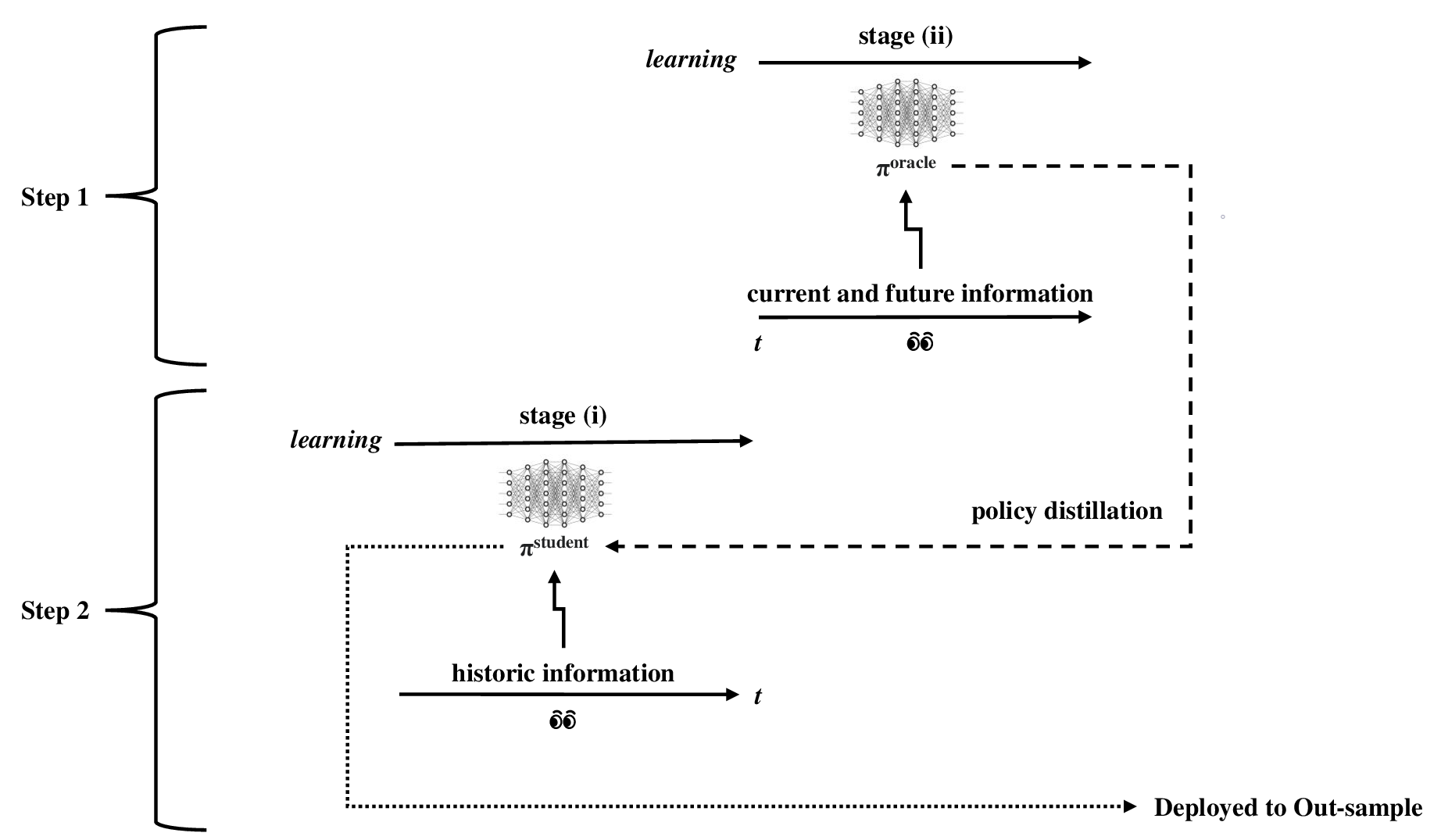}
    \caption{Training an oracle and student via IL}
    \label{fig:Training an oracle and student via IL}
\end{figure}

$ \phi $ is optimized by $\pi^*_\theta := \mathop{arg \ min}_{\theta } -\mathbb{E}( r \in \mathbfcal{R}^\phi \vert \mathbfcal{S}^\phi, \pi_\theta )$ where $\mathbfcal{R}$ is the accumulated reward of the trajectory, and $\mathbfcal{S}^\phi$ is the oracle’s state set which includes \textit{current and future data}, as previous data is irrelevant in optimizing for stage (ii). Here, $r \in \mathbfcal{R}^\phi$ is generated via the portfolio’s log return. The control task, $\pi^\phi: \mathbfcal{S}^\phi \longmapsto \mathbfcal{A}$ mapping can be optimized with any approach. The approaches can be largely divided into dynamical programming and learning-based approaches. Here we choose a learning-based MDP approach, namely actor-critic RL algorithms, Trust Region Policy Optimization (TRPO) \citep{schulman2015trust}, Proximal Policy Optimization (PPO) \citep{schulman2017proximal}, and Advantage Actor-Critic (A2C) \citep{mnih2016asynchronous}. These actor-critic approaches utilize both a policy (actor) and value (critic) network, which optimizes the state-action pair and state-value pair respectively, and finally combines the two into a single actor-critic algorithm. Neural-network-based MDP approaches are ideal for our problem as we attempt to identify a framework that can be applied universally across all financial sequential control tasks.

Due to $\mathbfcal{S}^\phi$, the optimal action at each time step $\pi^*(\mathbfcal{S}_t)$ is easily extractable. Policy optimization at each time step is trivial as $\phi$ has access to future values. Next, we distill the optimal policy via IL. Student $\psi$ is optimized by $\pi_\theta^\psi = \mathop{arg \ min}_{ \theta } -\mathbb{E}( r \in \mathbfcal{R}^\psi \vert \mathbfcal{S}^\psi, \pi_\theta )$, where $\mathbfcal{S}^\psi \equiv \mathbfcal{S}_{T'}$. The policy function $\pi^\psi: \mathbfcal{S}^\psi \longmapsto \mathbfcal{A}$ is far more challenging as $\mathbfcal{S}^\psi$ only has access to lagged data. I.e., it is responsible for stage (i) mentioned at the beginning of this section. Here, instead of setting the $r \in \mathbfcal{R}$ to the portfolio’s log return, $ r \in \mathbfcal{R}^\psi:= -distance(\mathbfcal{A}^\psi, \mathbfcal{A}^\phi = \mathbfcal{A}^*) $, in turn allowing $\pi^\psi$ to learn via the $\pi^\phi$ which theoretically reduces $\Delta noise$ in the label space. We set $distance(\cdot, \cdot)$ to the $\vert \vert \delta \vert \vert_2$ distance of the two vectors. As the learning algorithm aims to maximize the similarity of the two policy vectors—it equivalently aims to minimize the distance. Since $\pi^\phi$ and subsequently $\mathbfcal{A}^\phi$ is retrieved from offline data, $\mathbfcal{A}^\phi$ is only known as a posteriori---making it deterministic. The $\vert \vert \delta \vert \vert_2$ too is a deterministic mapping. Concretely, it would only be possible to retrieve a reliable stochastic $\pi^\phi$ if we could simulate parallel worlds---which we know is intractable.

\citep{fang2021universal}'s IL algorithm is named Oracle Policy Distillation (OPD), and it is trained via a linear combination of the return (the vanilla learner's reward) and the dissimilarity with the oracle. Our IL algorithm only uses the dissimilarity with the oracle, to examine a direct, pure form of IL. We hereforth refer to it as Direct Policy Distillation (DPD). The pseudo-code for the training process of IL approaches are available in Appendix Section 5.

\subsection{Curriculum Learning for Financial Control}
Our application of the CL paradigm takes an alternative approach in dealing with the $\Delta noise$. In the IL paradigm, the labels provided to the student is theoretically $\Delta noise$-free, however, this comes with removing a lot of the $\Delta signal$ in the label. This is consistent with our previous discussions mentioning that the per time step $\Delta signal$, and $\Delta noise$ is only decomposable as a posteriori and is, for practical purposes---intractable.

We implement the CL paradigm by smoothening the noisy time-series data during training. We hypothesize that minor data smoothing will reduce $\Delta noise$ more than $\Delta signal$, improving signal learning and out-sample performance. We attempt two types of data smoothing: (i) exponential moving averages (EMA) and (ii) rounding. Rounding is implemented as:

\[ \Delta sec_t^{rounded} \longleftarrow round(\Delta sec_t, 2) \tag{1}\]
\newline
where the second parameter represents the decimal places.

We prioritize EMA-based approaches because rounding is arbitrary, with less theoretical foundation than EMA. Our implementation of the EMA is defined recursively:

\[ \Delta sec_t^{EMA} \longleftarrow \alpha \cdot \Delta sec_t + (1 - \alpha) \cdot \Delta sec^{EMA}_{t-1} \tag{2} \]
\[ \alpha := 2/(w_l+1) \tag{3} \]
\newline
where $w_l$ is the lagging window hyperparameter. $\forall$EMA-based implementation, we concatenate $w_l$ to the end of the method name string.

The first approach we test is EMA5, where the training is done on EMA5 smoothed data. The second approach we test is Inverse-Smoothing (IS), directly inspired by CL. Given $S$, we let $w_l \longleftarrow S$ and train the network in $w_l$ distinct stages, with the first stage being the most smoothened, while the subsequent stage is $w_l \longleftarrow w_l - 1$. This results in the final stage having no smoothing as $w_l \longleftarrow 1$. Each stage is assigned an equal number of $\theta$ updates, as shown in Figure 3. The pseudo-code for the training process is available in Appendix Section 7.

\begin{figure}
    \centering
    \includegraphics[width=\columnwidth]{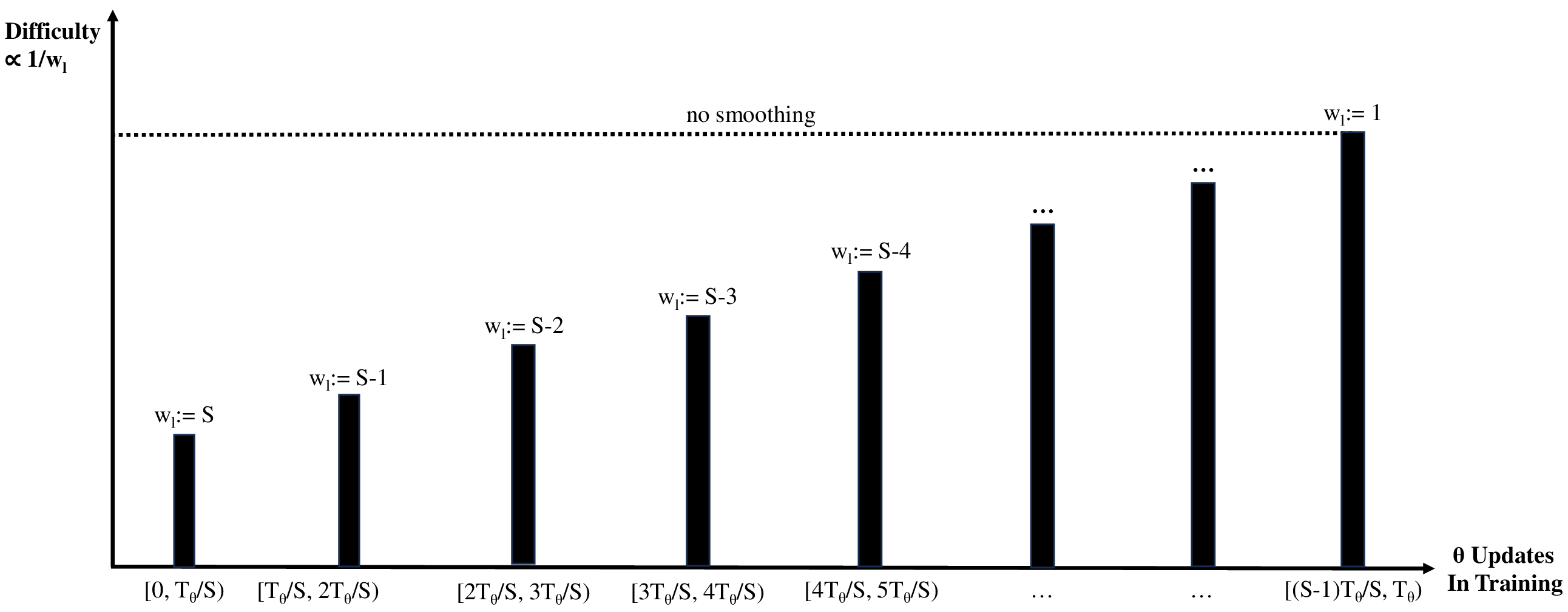}
    \caption{Inverse smoothing}
    \label{fig:Inverse Smoothing}
\end{figure}

\section{Data}

We select two representative data sets in the financial time series domain to test our ideas. The financial variables are chosen based on the greatest trading volume---i.e., how actively it is traded, which is a robust guage of its significance in the public financial markets. Notably, two data sets are used to examine (i) inter-asset-class and (ii) intra-asset-class effectiveness. That is, the first data set corresponds to a portfolio of different investment categories and financial variables, while the second data set corresponds to securities within a single investment category. We hereforth refer to Data Sets 1 and 2 as representing the Macro ETFs and Commodity Futures environments in Table 1, respectively. We visualize the stochastic time-series of $\mathbfcal{U}$ in Appendix Figures 6 and 7, with black vertical lines representing the train, validation, and test split. Further details on data sourcing and processing are provided in Appendix Section 1.

\section{Optimization Constraints}

To raise the robustness of our study, we apply two hard constraints, one for each data set---mimicking real-world portfolio optimization problems. Constraint 1 (Data Set 1): The portfolio's maximum aggregate gross exposure is 1. Equivalently, the net exposure is bound by [-1, 1]. Constraint 2 (Data Set 2): The portfolio's maximum aggregate gross exposure is 2. Equivalently, the net exposure is bound by [-2, 2]. We use simple linear normalization to respect the hard constraints. However, softmax can also be used.

\begin{align*}
\theta_{ML} = \mathop{arg \ min}_{ \theta } -\mathbb{E}_{(\mathbfcal{S}_t,r \in\mathbfcal{R}) \sim p_{data}}log \ p_{model}(r \vert \mathbfcal{S}_t ; \theta ) \\ \approx \hat{\theta}_{ML} = \mathop{arg \ min}_{ \theta } -\mathbb{E}_{(\mathbfcal{S}_t,r \in\mathbfcal{R}) \sim \hat{p}_{data}}log \ p_{model}(r \vert \mathbfcal{S}_t ; \theta ) \tag{4}
\end{align*}
\[ subject \ to \ -1 \leq \sum_{a \in \mathbfcal{A}_{\hat{\theta}}} w(a, \mathbfcal{A}_{\hat{\theta}}) \leq 1 \ \forall t \mathrm{, for \ Data \ Set \ 1},\]
\[ subject \ to \ -2 \leq \sum_{a \in \mathbfcal{A}_{\hat{\theta}}} w(a, \mathbfcal{A}_{\hat{\theta}}) \leq 2 \ \forall t \mathrm{, for \ Data \ Set \ 2}.\]
\newline
\section{Empirical Study}

By the Markov property, a non-sequential neural-network can be applied to the agent. However, to test the efficacy of a temporal encoder-decoder neural-network for the MDP environment, we examined some preliminary experiments with an LSTM architecture \citep{hochreiter1997long}. However, we found no improvement in performance but significantly slower training and inference run-times. Therefore, our experiments are Multilayer Perceptron (MLP)-based.

First, we compare our proposed approachs' performance against a heuristic and an RL baseline--- rebalanced portfolio (RP), and no IL and CL, respectively. RP is a common approach in financial control as this heuristic has been shown to beat most professional investment managers \citep{cover1991universal, ye2020reinforcement}. Additionally, OPD is a state-of-the-art baseline for IL-based control \citep{fang2021universal}. We constrain the OPD experiments to PPO as PPO is the only underlying algorithm tested by the authors of OPD. Additionally, we found it challenging to train OPD vis-à-vis run-time. That is, due to the higher computational complexity, we empirically observe and theoretically verify that the order of computational complexity is in descending order: MLP with OPD $>$ LSTM with all else $>$ MLP with all else. The remaining methods: \{DPD, DPD-LGN, R, EMA5, IS8, TIS\} are ours, where \{R\} is a na\"{\i}ve approach and \{DPD-LGN, TIS\} are ablation studies described in detail later in this section.

Each statistic in Tables 2, 3, and 4 is by nature highly stochastic, excluding RP, which is deterministic. $\forall$ RL-based methods, the statistics are mean$\pm 1\sigma$, where the sample size is 50. Notably, $\forall$inference, the model is trained with a random-seed. That is, each sample corresponds to an independently retrained model. This assures the robustness of our experiments. In aggregate, $2500 \longleftarrow (2 \cdot 8 \cdot 50 \cdot 3) + (2 \cdot 50)$ independent experiments have been conducted---(2 data sets, 8 stochastic methods, 50 samples, and 3 underlying model-free algorithms) + OPD on (2 data sets, 50 samples). We use the conventional 0.6, 0.2, and 0.2 split ratio for train, validation, and test set, respectively.

As we wish to examine the proposed methods' training efficiency and out-sample performance, we fix the RL step-count to $1,000,000$ steps of training, where the training set size $< 1,000,000$. We emphasize that this is unavoidable as the data set for $\hat{p}_{data}$ is fixed, and we are unable to further sample trajectories from ${p}_{data}$. This discussion is made in the previous sections. This leads to repetitive training of the same training set. To ensure that we are not in the overfitting regime, we conducted preliminary experiments examining the validation set performance at $[100,000:1,000,000]$ steps. Further details on hyperparameter tuning in the training, validation set, and inference on the test set are detailed in Appendix Sections 3, 4, 5, 6, 7, and 8.

\begin{table}
\centering
  \caption{Test Set Cumulative Return ($\pm1\sigma$): TRPO}
  \label{tab:commands}
  \scalebox{0.9}{
  \begin{tabular}{|c|ccc}
        \toprule
    Method & Data Set 1 & Data Set 2  \\
            \toprule
 \multicolumn{3}{c}{\textbf{Heuristic}} \\
  \toprule
     \rowcolor{Gray} RP & 2.951 & 91.057 \\
 \toprule
 \multicolumn{3}{c}{\textbf{Baseline}} \\
\toprule
    TRPO & 29.4402 $\pm$ 51.712 & $-12.820 \pm 38.931$  \\
\toprule
     \multicolumn{3}{c}{\textbf{Imitation Learning}*} \\
\toprule
   \rowcolor{Gray}
    TRPO-DPD & -1.762 $\pm$ 19.492 & -0.9014 $\pm$ 3.239 \\
    TRPO-DPD-LGN$^\dagger$ & -0.187 $\pm$ 17.948 & 1.849 $\pm$ 72.434 \\
\toprule
  \multicolumn{3}{c}{\textbf{Curriculum Learning}*} \\
\toprule
    \rowcolor{Gray}
    TRPO-R$^\ddagger$ & 21.949 $\pm$ 61.999 & -4.260 $\pm$ 49.437 \\
    TRPO-EMA5 & 104.599 $\pm$ 44.225 & 13.941 $\pm$ 67.672 \\   
    \rowcolor{Gray}   TRPO-IS8& 31.806 $\pm$ 29.014 & \textbf{49.471} $\pm$ \textbf{112.507} \\ 
    TRPO-TIS$^\dagger$ & \textbf{111.169} $\pm$ \textbf{44.182} & -56.721 $\pm$ 130.088 \\  
\bottomrule
\multicolumn{3}{c}{\small *ours, $^\dagger$ablation study, $^\ddagger$na\"{\i}ve approach} \\
  \end{tabular}}
\end{table}

\subsection{Ablation Study}

After observing the poor performances of IL approaches, and the improved performances of CL approaches, we found it imperative to further conduct \textit{two} additional ablation studies for a better understanding of the results.

\begin{table}
\centering
  \caption{Test Set Cumulative Return ($\pm1\sigma$): PPO}
  \label{tab:commands}
  \scalebox{0.95}{
  \begin{tabular}{|c|ccc}
        \toprule
    Method & Data Set 1 & Data Set 2  \\
            \toprule
 \multicolumn{3}{c}{\textbf{Heuristic}} \\
     \rowcolor{Gray}
        \toprule
    RP & 2.951 & 91.057 \\
   \toprule
 \multicolumn{3}{c}{\textbf{Baselines}} \\
   \toprule
    PPO & 29.440 $\pm$ 51.712 & 4.202 $\pm$ 64.185  \\
    \rowcolor{Gray}   PPO-OPD & 23.009 $\pm$ 48.459 & -44.461 $\pm$ 177.441 \\
\toprule
     \multicolumn{3}{c}{\textbf{Imitation Learning}*} \\
\toprule
   PPO-DPD & 3.543 $\pm$ 29.152 & -24.805 $\pm$ 79.796 \\
   \rowcolor{Gray}
   PPO-DPD-LGN$^\dagger$ & -3.059 $\pm$ 35.961 & -4.687 $\pm$ 100.238 \\
\toprule
  \multicolumn{3}{c}{\textbf{Curriculum Learning}*} \\
\toprule
    PPO-R$^\ddagger$ & 17.028 $\pm$ 50.973 & -13.728 $\pm$ 69.622 \\
    \rowcolor{Gray}
    PPO-EMA5 & 81.817 $\pm$ 33.423 & 35.215 $\pm$ 55.376 \\ 
    PPO-IS8 & \textbf{81.873} $\pm$ \textbf{39.850} & \textbf{39.389} $\pm$ \textbf{134.199} \\   
     \rowcolor{Gray}PPO-TIS$^\dagger$  & 28.032 $\pm$ 53.716 & 24.380 $\pm$ 120.121 \\

\bottomrule
\multicolumn{3}{c}{\small *ours, $^\dagger$ablation study, $^\ddagger$na\"{\i}ve approach} \\
  \end{tabular}}
\end{table}

First, synthetically removing the $\Delta noise$ in the label space, as seen in OPD and DPD, consistently resulted in poor performance. The transformation to a wholly deterministic label space in the process of policy distillation caused concern about the complete lack of $\Delta noise$. We know that deep learning models can benefit from perturbations \citep{liu2019universal, wong2020learning}, and wholly removing the noise may result in diminished generalization. Therefore, we implement Learned Gaussian Noise (LGN) on top of DPD, which tunes for the diagonal covariance matrix $\Sigma$ in the training and validation set. The mean vector is set to an all-zero vector to maintain the maximum likelihood properties of our optimization. With $\Sigma$, we perturb the label space. To efficiently optimize this continuous space, we use a large-scale Bayesian optimization implementation of \citep{JMLR:v21:18-223}. Further details are available in Appendix Section 6.

The second abalation study vis-à-vis CL tunes $S$, $S_{searchspace}:=\{1, 2,..., 8\}$ in the training and validation set, instead of $S:=8$. We name this approach as Tuned Inverse-Smoothing (TIS). We study whether tuning on the validation set is transferable out-sample, and the implications on the stationarity of $\Delta signal$ and $\Delta noise$ decomposition. Training for ablation studies are detailed in Appendix Sections 6 and 7.

\section{Analysis and Discussion}

\subsection{Results}

The results are summarized in Tables 2, 3, and 4 and visualized in Figures 4, 5, 8, 9, 10, and 11. Note that $\sigma$ in the Tables are derived by the entire test set, while the $\sigma$ in the Figures are per $t$. The heuristic RP achieves a 2.951\% and 91.057\% return in the test set. The visualization of the heuristic does not include $\pm \sigma$ as it is a deterministic policy. In the two environments, RL approaches display varying performance against the heuristic. This emphasizes the ongoing challenges of end-to-end neural-network-based control on highly stochastic time-series like financial markets. However, due to its deterministic nature, it is not possible to draw statistical conclusions that the heuristic will continue to perform well going forward. The heuristic is highly na\"{\i}ve, with no past information guiding future decision-making. 

\begin{table}
\centering
  \caption{Test Set Cumulative Return ($\pm1\sigma$): A2C}
  \label{tab:commands}
  \scalebox{0.9}{
  \begin{tabular}{|c|ccc}
        \toprule
    Method & Data Set 1 & Data Set 2  \\

            \toprule
 \multicolumn{3}{c}{\textbf{Heuristic}} \\
   \toprule
    \rowcolor{Gray}    RP & 2.951 & 91.057 \\
            \toprule
 \multicolumn{3}{c}{\textbf{Baseline}} \\
   \toprule
    A2C & 9.374 $\pm$ 33.604 & 9.567 $\pm$ 66.664 \\
\toprule
     \multicolumn{3}{c}{\textbf{Imitation Learning}*} \\
\toprule
   \rowcolor{Gray}
   A2C-DPD & 1.205 $\pm$ 40.249 & 0.099 $\pm$ 4.680 \\ 
   A2C-DPD-LGN$^\dagger$ & 9.613 $\pm$ 41.123 & 1.043 $\pm$ 15.055 \\
\toprule
  \multicolumn{3}{c}{\textbf{Curriculum Learning}*} \\
\toprule
    \rowcolor{Gray}
    A2C-R$^\ddagger$ & 7.493 $\pm$ 32.060 & 5.893 $\pm$ 72.870 \\
    A2C-EMA5 & 40.550 $\pm$ 24.504 & \textbf{104.17} $\pm$ \textbf{75.074} \\
    \rowcolor{Gray}
    A2C-IS8 & 46.018 $\pm$ 28.972 & 41.937 $\pm$ 106.385 \\   
    A2C-TIS$^\dagger$ & \textbf{46.414} $\pm$ \textbf{28.788} & 4.651 $\pm$ 130.721 \\

\bottomrule
\multicolumn{3}{c}{\small *ours, $^\dagger$ablation study, $^\ddagger$na\"{\i}ve approach} \\
  \end{tabular}}
\end{table}

Within the RL-based approaches, the best result across the column is highlighted in bold. We observe highly encouraging signs as $\forall$6 (2 data sets $\cdot$ 3 underlying algorithms) empirical environments, each best result is the EMA-based CL we propose---\{EMA5, IS8, TIS\}. Consistent with expectations, the na\"ive rounding approach's performance significantly lags the EMA-based approaches. Appendix Section 9, Tables 11, 12, and 13 presents a detailed statistical significance study.  

\begin{figure}
    \centering
    \includegraphics[width=\columnwidth]{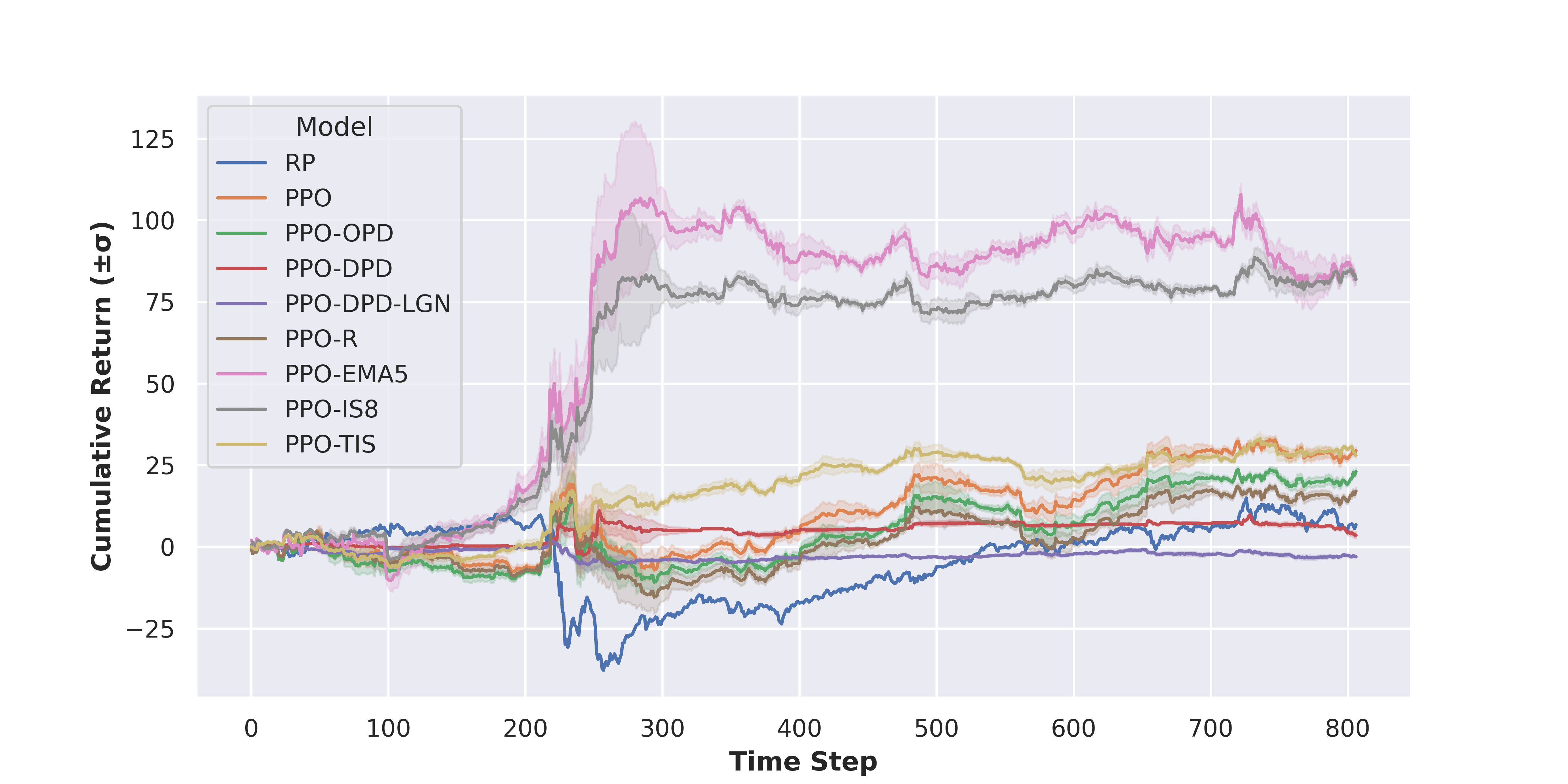}
    \caption{PPO Test Set Inference (Data Set 1)}
    \label{fig:Test Set Performance}
\end{figure}

Notably, out of the six empirical environments,  EMA5 achieves statistical significance ($<$ 0.05 P-value) all 6 times, while IS8 achieves statistical significance 5 times. The remaining non-statistically significant case still shows improvement over all baselines---with significant improvement against the heuristic. These results are highly encouraging as all overlapping hyperparameters, $h \in \mathbfcal{H}$ have been tuned for the baseline underlying algorithm, suggesting more room for improvement in our proposed methods on the training and validation stage.

\begin{figure}
    \centering
    \includegraphics[width=\columnwidth]{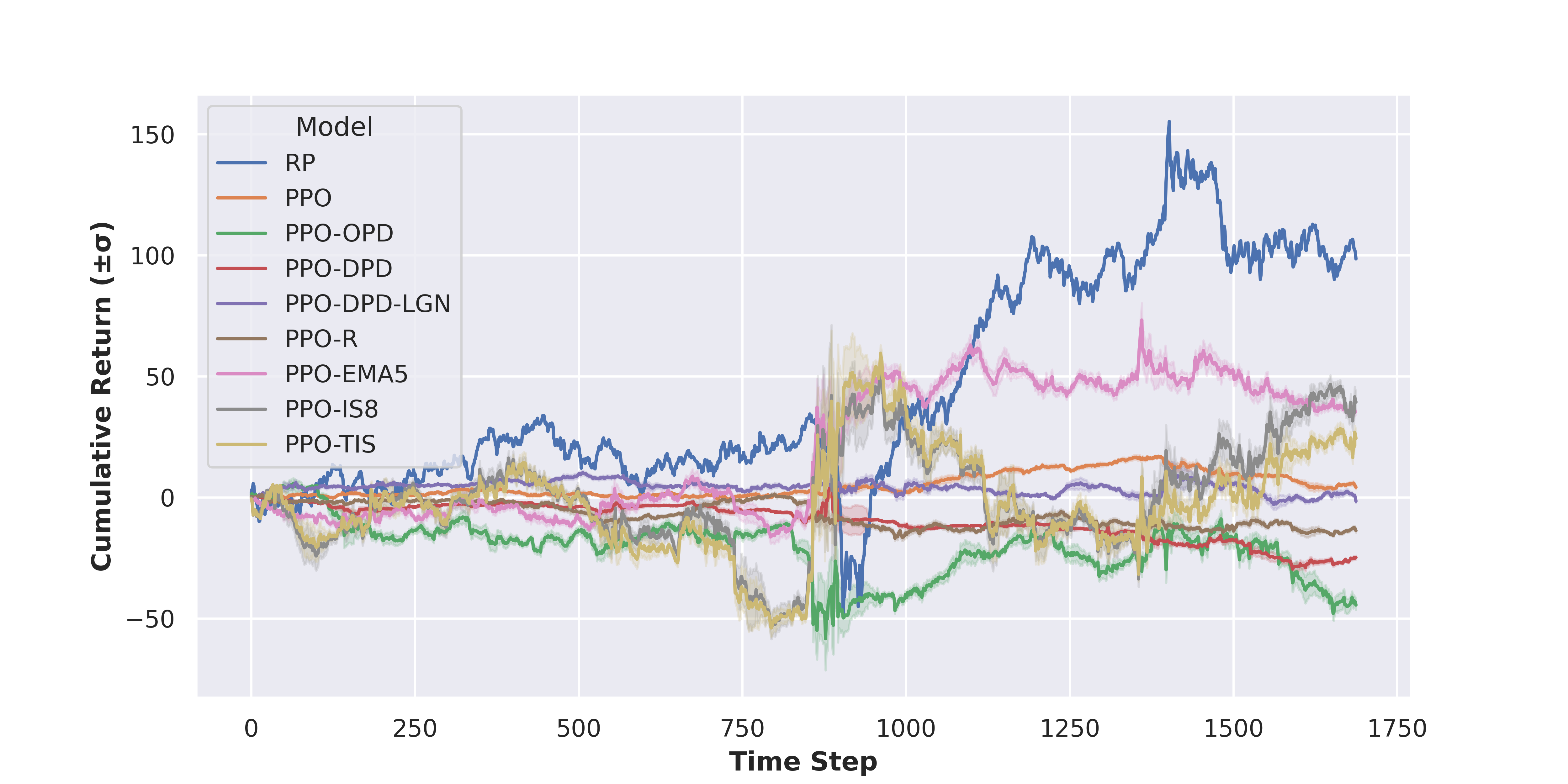}
    \caption{PPO Test Set Inference (Data Set 2)}
    \label{fig:Test Set Performance}
\end{figure}

On the other hand, we observe a dramatic worsening of performance for IL approaches. First, for the case of PPO-OPD, a state-of-the-art baseline, in this limited data environment, it distinctly performs worse than its underlying algorithm---PPO in both data sets. The DPD approach, which transforms the label space from stochastic to deterministic via policy distillation, fares no better. The added Gaussian noise in LGN occasionally improves performance, but no statistically meaningful observation is found. 

When IL approaches in Figures 4, 5, 8, 9, 10, 11 are examined in detail, it is clear that the agent has failed to learn any meaningful signal. The most extreme case can be viewed in Figure 11 where A2C-DPD and A2C-DPD-LGN refuses to take any position at all, resulting in a nearly flat test set performance. The performances of DPD, DPD-LGN, and OPD across all test set results make it obvious that minimal pattern recognition and learning has been achieved. We now analyze the results from the perspective of $\Delta signal$ and $\Delta noise$ decomposition.

\subsection{Theoretical Implications}
We begin with our first remark, which underlies the theoretical foundation for both the performance of CL and IL under the presented environment circumstances.

\begin{remark}
Given a possibly infinite or at least intractably large high dimensional underlying $p_{data}(\mathbfcal{A} \ \vert \ \mathbfcal{S}) \ \forall t$ due to the nature of determining $\mathbfcal{S}$, $\exists \Delta noise$ in both input and label space which adversely impacts deep reinforcement learning---which attempts to learn $\hat{p}_{data}(\mathbfcal{A} \ \vert \ \mathbfcal{S}_{approximate}) \ \forall t$.
\end{remark}

This remark allows us to define the second remark, which summarizes how the CL approaches improve deep reinforcement learning approaches under such circumstances. 

\begin{remark}
Given \textbf{\textup{Remark 1}}, it is possible to, on average, smooth the input and label space s.t. 

\[
\ \frac{1}{\vert D_T \vert}
\sum_{d \in D_T}{\frac{\partial P_G}{-(\partial \Delta noise_d)} - \frac{\partial P_G}{-(\partial \Delta signal_d)}} 
> 0
\tag{5}
\]
\newline
where smoothing necessarily $\partial \Delta noise \leq 0$ and $\partial \Delta signal \leq 0$. $D_T$ refers to the set of data points for training, and $P_G$ refers to generalization performance, i.e., out-sample performance. 
\end{remark}

For reasons described in previous sections, $\Delta signal$ and $\Delta noise$ are intractable. Therefore, we are left with indiscriminate smoothing that often leads to $\partial \Delta noise < 0 \implies \partial \Delta signal < 0$, $\forall \frac{\partial P_G}{-(\partial \Delta noise)} > 0$. Notably, our empirical study supports in-so-far the ``on average" clause, as the entire training data, including the input and label space, is smoothed identically. Despite the overwhelming evidence supporting the direct use of CL-based smoothing approaches on noisy control tasks, hyperparameter tuning the degree of smoothing can be tricky, as we observe that it shows signs of the non-stationary property.

\begin{remark}
The decomposition $\Delta signal + \Delta noise \longleftarrow \Delta sec$ is likely non-stationary. I.e., $\frac{\Delta signal}{\Delta sec}\sim p_s$ and $\frac{\Delta noise}{\Delta sec} \sim p_n$ changes over the temporal dimension.
\end{remark}

We observe this empirically, as TIS does not always lead to the best performance. We note that the tuning only employs 1 sample per $S \in S_{searchspace}$, making it a biased estimate of the optimal hyperparameter. This is why we mention that the non-stationary property is likely in Remark 3. The claim is not statistically significant. Despite this shortcoming, we observe significant instability in the model's ability to generalize when employing the tuned hyperparameter given in Appendix Table 8. Notably, it identified that $S \longleftarrow 1$ is optimal in two of the six cases, corresponding to no CL. However, consistent with the superior performance of smoothed training data points, its performance falls dramatically in these cases. Therefore, we leave it up to future work to investigate practical workarounds to CL's non-stationary nature of the smoothing hyperparameter.

The poor performance of IL approaches under these circumstances can be described by Remark 4.

\begin{remark}
Given \textbf{\textup{Remark 1}}, $\Delta noise$ can be synthetically removed in the label space during training via policy distillation from an oracle. However, this dramatic reduction in $\Delta noise$ transforms the label space from stochastic $\longrightarrow$ deterministic, losing most of the $\Delta signal$ in the process.
\end{remark}

The mechanism of Remark 4 is carefully detailed in Appendix Section 10, which we urge the reader to examine. We believe this theoretical analysis is helpful for future research in IL approaches when the underlying system is very high dimensional---and, in turn, stochastic.

\section{Future Works}

This work opens the door to an extensive list of possible future works. The most pressing next step is to examine the non-stationary nature of the signal, noise decomposition. The results here hint at a possibly non-stationary decomposition---meaning robust CL approaches will benefit from dynamically adapting the level of smoothing over the temporal dimension. Furthermore, only a few smoothing methods have been empirically tested here. We encourage future works to attempt other smoothing methods and analyze the theoretical reasoning behind the smoothing technique.

Additionally, future works can branch out from the specific environment presented here. First, investigating other highly stochastic temporal systems and domains that may benefit from the proposed approach will be impactful. Moreover, there is no theoretical reason why the proposed CL approaches would not transfer to other learning tasks, such as forecasting.

\par\vspace{\fill}\pagebreak[0]

\section{Appendix}

\subsection{1. Data Sourcing and Processing}

We obtain both data sets from S$\&$P Capital IQ and Bloomberg to ensure the quality and consistency of the data, and to prevent look-ahead bias. We source the Bull-Bear Spread separately from the Investor Sentiment Index of the American Association of Individual Investors (AAII). All the data used in the empirical study is available publically and have been extensively used as benchmarks in representative studies \citep{dixon2015implementing,deng2016deep,gudelek2017deep,sezer2018algorithmic,pendharkar2018trading}. 

The financial variables composing the data sets are available in Table 5 and 6. In Data Set 1, corresponding to Table 5, $\ \mathbfcal{U}_1=\{GLD, USO, USD, LQD\}$ and $\mathbfcal{M}_1=\{3M, 2Y, 10Y, FFEOR, 10Y-3M, 10Y-2Y, IYR, VIX, BULL\_BEAR\_SPREAD\}$. The initial time step is set to the date where $\exists$  valid data points $\forall$ variable. The Data Set 1's date spans: [2006-04-11, 2022-07-08] in daily units. Data Set 2's date spans: [1990-01-021, 2023-06-26] in daily units. $\mathbfcal{U}$ is visualized in Figure 6 and 7, with black vertical lines representing the train, validation, and test split.

\begin{table} [!htb]
  \caption{Data Set 1 (Inter-Asset-Class)}
  \label{tab:freq}

  \resizebox{\columnwidth}{!}{
    \begin{tabular}{ccc} 
    \toprule
    Asset-Class&Variable&Abbreviation\\
    \midrule
    \rowcolor{Gray}Commodity& SPDR Gold Trust & GLD\\
   Commodity&U.S. Oil Fund & USO\\
    \rowcolor{Gray}Currency&U.S. Dollar Index & USD\\
Fixed Income&U.S. IG Corporate Bond & LQD\\
\rowcolor{Gray}Interest Rate&3M Treasury Yield & 3M\\
Interest Rate&2Y Treasury Yield & 2Y\\
\rowcolor{Gray}Interest Rate&10Y Treasury Yield & 10Y \\
Interest Rate&Fed Funds Effective Rate & FFEOR\\
\rowcolor{Gray}Rate Spread&10Y-3M Spread & 10Y-3M\\
Rate Spread&10Y-2Y Spread & 10Y-2Y\\
\rowcolor{Gray}Real Estate&U.S. Real Estate & IYR\\
Risk&CBOE Volatility Index  & VIX\\
\rowcolor{Gray}Sentiment&Bull-Bear Spread  & BULL\_BEAR\_SPREAD\\
  \bottomrule
\end{tabular}
  }
\end{table}

\begin{table} [!htb]
  \caption{Data Set 2 (Intra-Asset-Class)}
  \label{tab:freq}
\centering
    \begin{tabular}{ccl} 
    \toprule
    Asset-Class&Variable\\
    \midrule
    \rowcolor{Gray}Commodity &Wheat Futures \\
    Commodity &Corn Futures\\
          \rowcolor{Gray}Commodity &Copper Futures\\
   Commodity &Silver Futures\\
    \rowcolor{Gray}Commodity &Gold Futures \\
        Commodity &Platinum Futures \\
    \rowcolor{Gray}Commodity &Crude Oil Future\\
         Commodity &Heating Oil Futures\\
  \bottomrule
\end{tabular}
\end{table}

The only data processing performed on the raw data is converting price data to return data and pre-processing any missing values. We use log returns for market variables, a standard practice in the financial domain. Log returns are used instead of regular differences because they result in more convenient downstream computations. Other data points are converted to regular differences because their values are much smaller and require higher precision. Market variables (that are transformed to log return, LR) for Data Set 1 is $\textbf{LR}_1:= \{GLD, USO, USD, LQD, IYR, VIX\}$, while for Data Set 2 is $\textbf{LR}_2 \equiv \mathbfcal{U}_2$, which are all the variables listed in Table 6. Since $u \in \mathbfcal{U}$ are tradable---i.e., can be interacted with via the market, they are market variables by definition. Therefore, $\mathbfcal{U} \subseteq \textbf{LR}$. In Data Set 1, $\exists$ non market variables $\textbf{LR}'_1= \{3M, 2Y, 10Y, FFEOR, 10Y-3M, 10Y-2Y, BULL BEAR SPREAD\}$. This distinction is useful for the data processing and the state representation method we present in Appendix Section 2. The pseudo-code for the data processing is provided in Algorithm 1.

\begin{figure}
    \centering
    \includegraphics[width=\columnwidth]{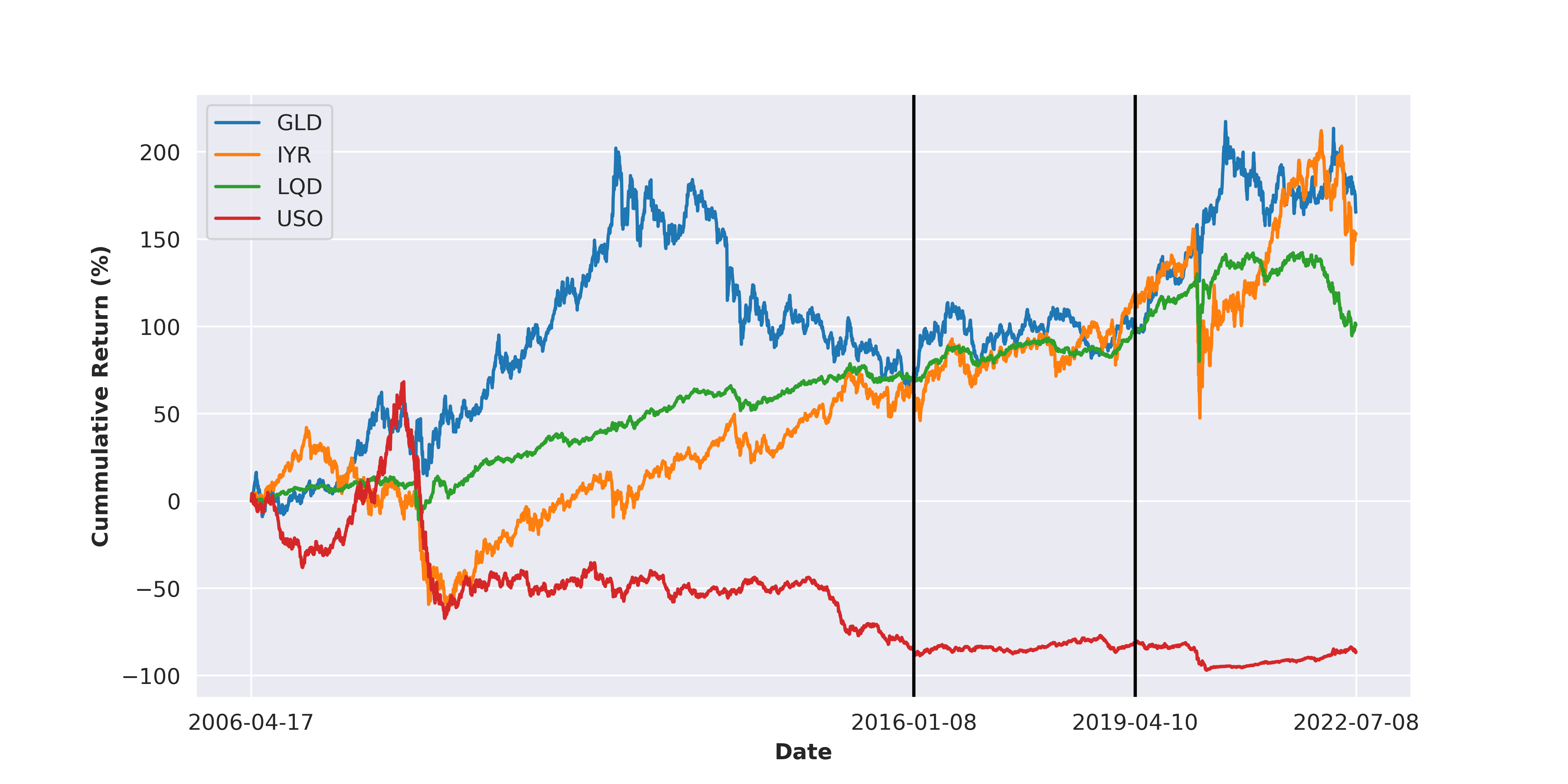}
    \caption{$\mathbfcal{U}$ of Macro ETFs (Data Set 1)}
    \label{fig:Macro ETFs}
\end{figure}

\begin{figure}
    \centering
    \includegraphics[width=\columnwidth]{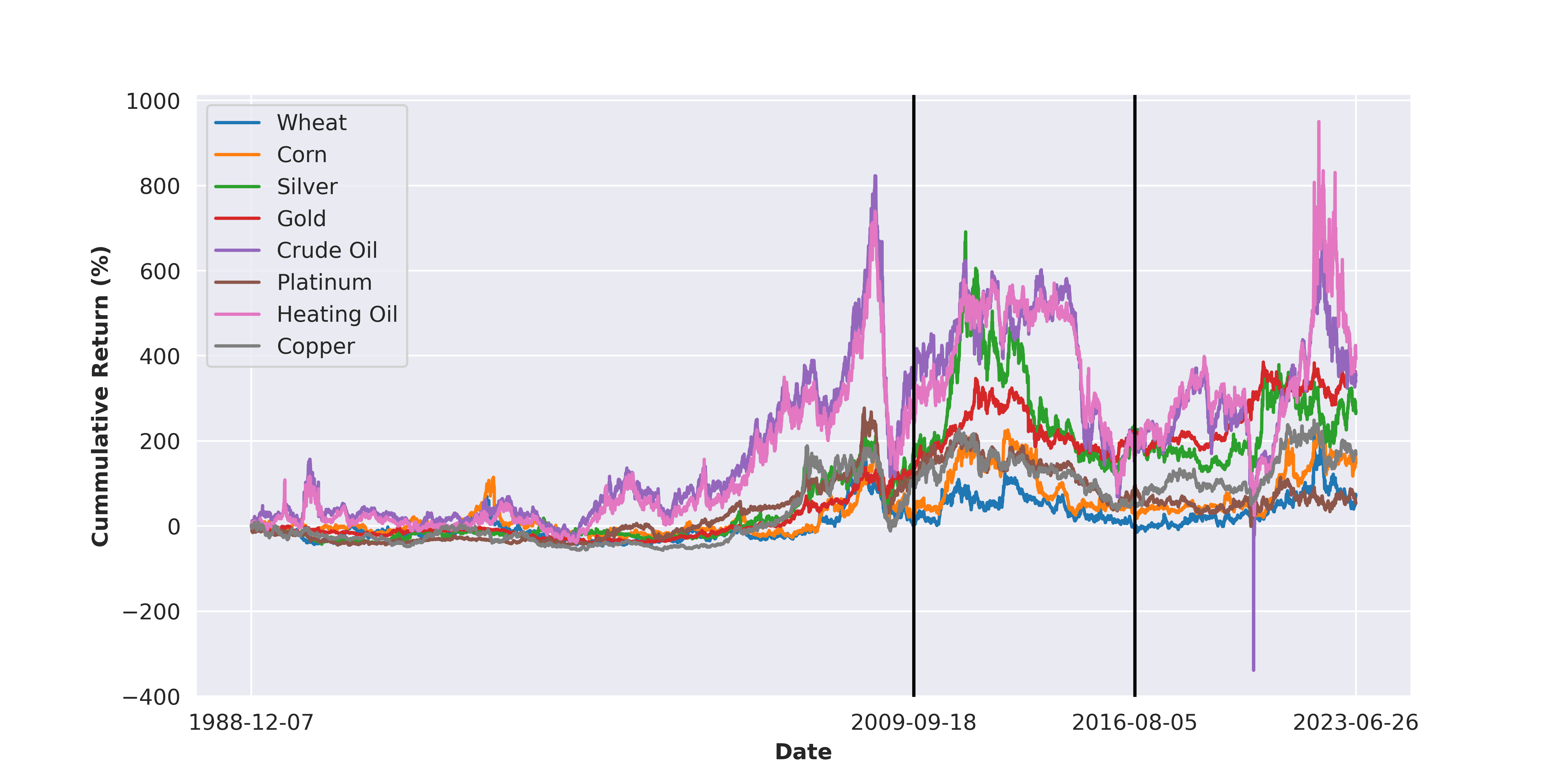}
    \caption{$\mathbfcal{U}$ of Commodity Futures (Data Set 2)}
    \label{fig:Commodity Futures}
\end{figure}

\begin{algorithm*}[!htb]
\textbf{Input}: $data_{raw}$\\
\textbf{Output}: $data_{processed}$

\begin{algorithmic}[0] 
\STATE $init \ \ data_{processed}$
\STATE $\textbf{F} \longleftarrow data_{raw}.get\_feature\_set()$\
\STATE \;
\FOR {$f \in \textbf{F}$}
\IF {$f$ \textbf{is} $MarketVariable$}
\STATE $\forall data_{processed}[f].datapoint[i] \longleftarrow log(data_{raw}[f].datapoint[i]/data_{raw}[f].datapoint[i-1])$
\STATE \;
\ELSIF {$f.datapoints \neq nan$}
\STATE $\forall data_{processed}[f].datapoint[i] \longleftarrow data_{raw}[f].datapoint[i]-data_{raw}[f].datapoint[i-1]$
\STATE \;
\ELSE
\STATE $\forall data_{processed}[f].datapoint[i] \longleftarrow data_{raw}[f].datapoint[i]-data_{raw}[f].datapoint[most\_recent\_non\_nan\_i < i]$
\ENDIF
\ENDFOR
\STATE \;
\STATE $data_{processed}[\textbf{F}].datapoint[0].drop\_timestep()$\;
\RETURN $data_{processed}$
\end{algorithmic}

\caption{Data Process}
\label{Data Process}
\end{algorithm*}

\subsection{2. State Representation}
By the MDP framework, the decision-making $\forall t$ can be made only given $\mathbfcal{S}_t$. That is, all necessary information for optimizing $\pi_t$ should be available in $\mathbfcal{S}_t$. Various methods exist to generate a state representation that captures historical data points. In preliminary tests, we examine an LSTM-based context vector for summarizing historical data points $\mathbfcal{S}_t$; however, we find no meaningful improvement in control performance with a rise in computational run-time. We found a simple heuristic representation where $T'$ is tuned in the training and validation set, which was computationally resource-efficient and worked consistently well. Hyperparameter $T'$ corresponds to State Lag in Tables 9 and 10. 

The search space for $T'$ was $T'_{searchspace}:= \{5, 10,..., 55, 60\}$ where, e.g., $T':=5$ would mean that information on the last five data points are incorporated in $\mathbfcal{S}$. However, to avoid very large $\vert \mathbfcal{S} \vert$, we include the past 5 cumulative returns up to $t-5$, but after that, every increase of 5 adds one additional element to $\mathbfcal{S}$. E.g., $T':=20 \Rightarrow \vert \mathbfcal{S}_{T'} \vert = 5 + (1\cdot 3) = 8$.
This is better understood via the state representation generating pseudo-code in Algorithm 2.

\begin{algorithm*}[!htb]
\caption{State Representation}
\label{alg:algorithm}
\textbf{Input}: $data_{processed}$, $\textbf{LR}$, $\textbf{LR}'$, $T'$ \\
\textbf{Output}: $\mathbfcal{S}_t$
\begin{algorithmic}[1] 

\STATE $init \ \ vector \ \ \mathbfcal{S}_t$, $state_{LR}$, $state_{LR'}$
\STATE $state_{LR'} \longleftarrow data_{processed}[t-1][\textbf{LR}']$
\STATE 
\FOR{$feature \in \textbf{LR}$ }
\STATE $init \ \ vector \ \ temp$
\STATE $lag_{size} \longleftarrow 5 + (T'-5)/5$
\STATE
\FOR{$i \in \{0, 1,.., lag_{size}\}$}

\IF{$i < 6$}
\STATE $temp.append(sum(data_{processed}[t-1-i:t-1[feature]))$
\ELSE
\STATE $i \longleftarrow (i - 5) * 5 + 5$
\STATE $temp.append(sum(data_{processed}[t-1-i:t-1][feature]))$
\ENDIF

\ENDFOR
\STATE

\STATE $state_{LR}.append(temp)$
\STATE $state_{LR} \longleftarrow Reshape(state_{LR}, (-1))$

\ENDFOR

\STATE
\STATE $\mathbfcal{S}_t \longleftarrow Concat(state_{LR'}, state_{LR})$

\STATE \textbf{return} $\mathbfcal{S}_t$
\end{algorithmic}
\end{algorithm*}

\subsection{3. Hyperparameters}

Notably, for a fair comparison, we keep the hyperparameters overlapping across underlying algorithms and methods fixed to compare them. Concretely, $\exists$ set of overlapping hyperparameters, $\mathbfcal{H}$ $\forall$ (Data Set, Underlying RL Algorithm) pair, therefore in aggregate $\vert \mathbfcal{H} \vert = 6 \longleftarrow 2 \cdot 3$ are used here. $\forall h \in \mathbfcal{H}$ is tuned by training on the train set and optimizing on the validation set. The optimized hyperparameters are available in Tables 7, 8, 9, and 10. One additional hyperparameter exists for the OPD method---distillation loss coefficient, which was tuned to $0.5$. We use random grid search and Bayesian optimization for hyperparameter tuning---with implementation details in Appendix sections 3, 4, 5, and 6.

\begin{table} [!htb]
\centering
  \caption{Learned Gaussian Noise: $\Sigma$}
  \label{tab:commands}
  \scalebox{0.9}{
  \begin{tabular}{|c|ccc}
        \toprule
    Method & Data Set 1 & Data Set 2  \\
            \toprule
    \rowcolor{Gray}
    $\sigma$ & 1.628 & 2.170 \\

\bottomrule
  \end{tabular}}
\end{table}

\begin{table} [!htb]
\centering
  \caption{Tuned Inverse Smoothing Hyperparameter: S}
  \label{tab:commands}
  \scalebox{0.9}{
  \begin{tabular}{|c|ccc}
        \toprule
    Method & Data Set 1 & Data Set 2  \\
            \toprule
   
    \rowcolor{Gray}
    TRPO-TIS & 8 & 1 \\
    PPO-TIS & 3 & 7 \\
     \rowcolor{Gray}   A2C-TIS & 8 & 1 \\
\bottomrule
  \end{tabular}}
\end{table}

\begin{table}[!htb]
\centering
  \caption{Tuned $h \in \mathbfcal{H}$ for Data Set 1}
  \label{tab:commands}
  \scalebox{0.9}{
  \begin{tabular}{|c|ccc}
        \toprule
    $h$ & TRPO & PPO & A2C \\
    \toprule
    \rowcolor{Gray}Learning Rate & 0.0001&0.0001  & 0.0001 \\
 State Lag & 10 & 10 & 10 \\
    \rowcolor{Gray}Steps per Update  & 292 & 292 & 292  \\
 Partition Factor for Batch Size  & 4 &4 &--- \\
    \rowcolor{Gray}Epochs & --- & 8& --- \\
Discount Factor     & 0.956 & 0.956 & 0.956  \\
    \rowcolor{Gray}Bias-Variance Trade-off Factor& 0.94 & 0.94 & 0.94 \\
 Clipping Parameter& --- & 0.6 & --- \\
    \rowcolor{Gray}Entropy Coefficient& --- & 0.03 & 0.03  \\
Value Function Coefficient& --- & 4.6 & 4.6 \\
\rowcolor{Gray}Conjugate Gradient Max Steps&  15 &---&---  \\
Hessian Dampening & 0.1  & ---&--- \\
\rowcolor{Gray}Line Search Reduction Factor& 0.8 &---&---  \\
Line Search Maximum Iteration& 10  &---&---  \\
\rowcolor{Gray}Critic Updates per Policy Update&  10 &---&---  \\
Target KL Divergence& 0.01 &---&---  \\
\rowcolor{Gray}Sub-sampling Factor& 1 &---&---  \\
Max Gradient Clipping& --- &--- & 0.6 \\
\rowcolor{Gray}RMSProp Epsilon& --- &--- & 0.0 \\

\bottomrule
  \end{tabular}}
\end{table}

\begin{table}[!htb]
\centering
  \caption{Tuned $h \in \mathbfcal{H}$ for Data Set 2}
  \label{tab:commands}
  \scalebox{0.9}{
  \begin{tabular}{|c|ccc}
        \toprule
    $h$ & TRPO & PPO & A2C \\
    \toprule
    \rowcolor{Gray}Learning Rate &  0.01 & 0.01 &0.01 \\
 State Lag &  45 & 45 & 45\\
    \rowcolor{Gray}Steps per Update  & 73 & 73 & 73 \\
 Partition Factor for Batch Size  & 2 & 2 & --- \\
    \rowcolor{Gray}Epochs & --- & 11& --- \\
Discount Factor     & 0.908 & 0.908 & 0.908  \\
    \rowcolor{Gray}Bias-Variance Trade-off Factor& 0.94 & 0.94 & 0.94  \\
 Clipping Parameter& --- & 0.35& --- \\
    \rowcolor{Gray}Entropy Coefficient& --- & 0.02 & 0.02 \\
Value Function Coefficient& --- &0.5& 0.5 \\
\rowcolor{Gray}Conjugate Gradient Max Steps& 15 &---&---  \\
Hessian Dampening &  0.1& ---&--- \\
\rowcolor{Gray}Line Search Reduction Factor& 0.8 &---&---  \\
Line Search Maximum Iteration& 10 &---&---  \\
\rowcolor{Gray}Critic Updates per Policy Update& 10 &---&---  \\
Target KL Divergence& 0.01 &---&---  \\
\rowcolor{Gray}Sub-sampling Factor& 1 &---&---  \\
Max Gradient Clipping& --- &--- & 0.6  \\
\rowcolor{Gray}RMSProp Epsilon& --- &--- & 0.0  \\

\bottomrule
  \end{tabular}}
\end{table}

\subsection{4. Training Baselines}

RP is a heuristic-based deterministic baseline and does not need training. On the contrary, TRPO, PPO, and A2C need training. The pseudo-code is presented in Algorithm 3. The subscript in $TrainModel()$ is the number of MDP steps the neural-network is trained on. $tune\_sample\_count:=150$ and $tune\_step\_size:=100,000$ for all methods, including IL and CL. All training for the models presented in this paper is done on machines up to a single RTX 3090, 16 CPU cores (32 threads), and 32 GB of RAM. This means that computationally weaker units have also been used to train numerous experiments in parallel. 

\begin{algorithm*}[!htb]
\caption{TRPO, PPO, A2C Training}
\label{alg:algorithm}
\textbf{Input}: $data_{processed}$, $\mathbfcal{H}_{searchspace}$, $split\_ratio, test\_sample\_count, tune\_sample\_count, tune\_step\_size$ \\
\textbf{Output}: $\boldsymbol{\pi}_{trained}, \mathbfcal{H}$
\begin{algorithmic}[1] 
\STATE $data_{train}, data_{validation}, data_{test}\longleftarrow split\_ratio(data_{processed})$
\STATE

\STATE $samples \longleftarrow RandomGridSearch(\mathbfcal{H}_{searchspace}, data_{train}, data_{validation}, tune\_step\_size, $

$tune\_sample\_count)$

\STATE $\mathbfcal{H}^* \longleftarrow arg\_max(samples.r \in \mathbfcal{R}_{validation})$
\STATE
\STATE $data_{train} \longleftarrow Concat(data_{train}, data_{validation})$
\STATE

\FOR{$\{0, 1, ..., test\_sample\_count-1\}$}
\STATE $\pi_{trained} \longleftarrow TrainModel_{1,000,000}(data_{train}, \mathbfcal{H}^*, AdamOptimizer)$
\STATE $\boldsymbol{\pi}_{trained}.append(\pi_{trained})$
\ENDFOR
\STATE \textbf{return} $\boldsymbol{\pi}_{trained}, \mathbfcal{H}^*$
\end{algorithmic}
\end{algorithm*}

\subsection{5. Imitation Learning Training}
The pseudo-code for IL methods is presented in Algorithm 4. The original work \citep{fang2021universal} can be reviewed for a detailed implementation of OPD. The DPD framework is presented in the main body.

Notably, $\phi$ does not require the $state\_lag$ hyperparameter as historical values are useless for an oracle (future-seeing). Also, $\phi$ is trained for 5,000,000 steps to ensure a global optima. In DPD, lines 6 and 7 corresponding to finding a global optima, can be achieved via any method---including analytical. We identify that 5,000,000 steps are more than enough to find a global optima using RL. Since the global optima of historical data is deterministic, we only need to train the $\phi$ once to extract the optimal trajectory.

\begin{algorithm*}[!htb]
\caption{OPD, DPD Training}
\label{alg:algorithm}
\textbf{Input}: $data_{processed}$, $\mathbfcal{H}_{searchspace}^{IL}$, $\mathbfcal{H}^*$, $split\_ratio$, $test\_sample\_count$, $tune\_sample\_count$, $tune\_step\_size$ \\
\textbf{Output}: $\boldsymbol{\pi}_{trained}$, $\mathbfcal{H}^{\psi*}$
\begin{algorithmic}[1] 
\STATE $data_{train}, data_{validation}, data_{test}\longleftarrow split\_ratio(data_{processed})$
\STATE

\STATE \begin{flushleft} $samples^\phi \longleftarrow RandomGridSearch^\phi(\mathbfcal{H}_{searchspace}^{IL}, \mathbfcal{H}^*\backslash \{state\_lag\} , data_{train}, data_{validation},$ \end{flushleft}
$tune\_step\_size, tune\_sample\_count)$

\STATE $\mathbfcal{H}^{\phi*} \longleftarrow arg\_max(samples^\phi.r \in \mathbfcal{R}_{validation})$

\STATE
\STATE $\pi^\phi \longleftarrow TrainModel_{5,000,000}(data_{train}, data_{validation}, \mathbfcal{H}^{\phi*}, AdamOptimizer)$
\STATE

\STATE \begin{flushleft}$samples^\psi \longleftarrow RandomGridSearch^\psi(\pi^\phi, \mathbfcal{H}_{searchspace}^{IL}, \mathbfcal{H}^*, data_{train}, data_{validation},$
$tune\_step\_size, tune\_sample\_count)$\end{flushleft}

\STATE $\mathbfcal{H}^{\psi*} \longleftarrow arg\_max(samples^\psi.r \in \mathbfcal{R}_{validation})$
\STATE
\STATE $data_{train} \longleftarrow Concat(data_{train}, data_{validation})$
\STATE
\FOR{$\{0, 1, ..., test\_sample\_count-1\}$}
\IF{OPD}
\STATE $\pi^\psi_{trained} \longleftarrow ImitationTrainModel^{OPD}_{1,000,000}(\pi^\phi, data_{train}, \mathbfcal{H}^{\psi*}, \mathbfcal{H}^*, AdamOptimizer)$
\ELSE
\STATE $\pi^\psi_{trained} \longleftarrow ImitationTrainModel^{DPD}_{1,000,000}(\pi^\phi, data_{train}, \mathbfcal{H}^{\psi*}, \mathbfcal{H}^*, AdamOptimizer)$
\ENDIF
\STATE $\boldsymbol{\pi}_{trained}.append(\pi^\psi_{trained})$
\ENDFOR
\STATE \textbf{return} $\boldsymbol{\pi}_{trained}$, $\mathbfcal{H}^{\psi*}$, $\pi^\phi$
\end{algorithmic}
\end{algorithm*}

\subsection{6. Imitation Learning Ablation Training}
DPD-LGN training is presented in Algorithm 5. The only added change is the label space for the student---the co-domain of $\pi^\phi$, which has now been perturbed via a Gaussian process. \citep{JMLR:v21:18-223}'s parallel Bayesian optimizer is used with a bandit optimizer and Euclidean distance.

\begin{algorithm*}[!htb]
\caption{DPD-LGN Training}
\label{alg:algorithm}
\textbf{Input}: $data_{processed}$, $\mathbfcal{H}^{\psi*}$, $\mathbfcal{H}^*$, $\pi^\phi$, $\sigma_{searchspace}$, $split\_ratio$, $test\_sample\_count$, 
$tune\_sample\_count$, $tune\_step\_size$ \\
\textbf{Output}: $\boldsymbol{\pi}_{trained}$
\begin{algorithmic}[1] 
\STATE $data_{train}, data_{validation}, data_{test}\longleftarrow split\_ratio(data_{processed})$

\STATE
\STATE \begin{flushleft} $\Sigma^*\longleftarrow$ $ParallelBayesianOptimizer(\sigma_{searchspace}$, $\mathbfcal{H}^{\psi*}$, $\mathbfcal{H}^*$, $\pi^\phi$, $data_{train}$, $data_{validation}$, $tune\_step\_size$, $tune\_sample\_count)$\end{flushleft}
\STATE $\pi^\phi \longleftarrow GaussianPerturbation(\pi^\phi, \mu:=0,\Sigma:=\Sigma^*)$
\STATE
\STATE $data_{train} \longleftarrow Concat(data_{train}, data_{validation})$

\STATE
\FOR{$\{0, 1, ..., test\_sample\_count-1\}$}
\STATE $\pi^\psi_{trained} \longleftarrow ImitationTrainModel_{1,000,000}(\pi^\phi, data_{train}, hyperparameters, AdamOptimizer)$
\STATE $\boldsymbol{\pi}_{trained}.append(\pi^\psi_{trained})$
\ENDFOR
\STATE \textbf{return} $\boldsymbol{\pi}_{trained}$
\end{algorithmic}
\end{algorithm*}

\subsection{7. Curriculum Learning and Ablation Training}

All methods presented under CL apply some form of data augmentation only during training. Inference on the validation and test set are never augmented to preserve maximum likelihood properties in estimating out-sample $p_{data}$. Notably, during training, both the state space $\mathbfcal{S}$ and label space $r \in \mathbfcal{R}$ are smoothed. R and EMA are presented in Algorithm 6, while IS and TIS are presented in Algorithm 7 and 8. $S_{searchspace}:=\{0, 1, ...,8\}$ for TIS, while $S_{searchspace}:=\{\#\}$ for IS\#, where $\# \in \mathbb{N}$ of choice. In our case, IS8 corresponds to $S_{searchspace}:=\{8\}$.

\begin{algorithm*}[!htb]
\caption{R, EMA Training}
\label{alg:algorithm}
\textbf{Input}: $data_{processed}$, $\mathbfcal{H}^*$, $split\_ratio$, $test\_sample\_count$, $tune\_sample\_count$, $tune\_step\_size$, $rounding\_parameter$ or $w_l$ \\
\textbf{Output}: $\boldsymbol{\pi}_{trained}$
\begin{algorithmic}[1] 
\STATE $data_{train}, data_{validation}, data_{test}\longleftarrow split\_ratio(data_{processed})$
\STATE $data_{train} \longleftarrow Concat(data_{train}, data_{validation})$
\STATE $data_{train} \longleftarrow Smoothing(data_{train}, rounding\_parameter \ or \ w_l)$
\STATE

\FOR{$\{0, 1, ..., test\_sample\_count-1\}$}
\STATE $\pi_{trained} \longleftarrow TrainModel_{1,000,000}(data_{train}, \mathbfcal{H}^*, AdamOptimizer)$
\STATE $\boldsymbol{\pi}_{trained}.append(\pi_{trained})$
\ENDFOR
\STATE \textbf{return} $\boldsymbol{\pi}_{trained}$
\end{algorithmic}
\end{algorithm*}

\begin{algorithm*}[!htb]
\caption{IS, TIS Training}
\label{alg:algorithm}
\textbf{Input}: $data_{processed}$, $\mathbfcal{H}^*$, $split\_ratio, test\_sample\_count, tune\_sample\_count, tune\_step\_size$, $S_{searchspace}$ \\
\textbf{Output}: $\boldsymbol{\pi}_{trained}$
\begin{algorithmic}[1] 
\STATE $data_{train}, data_{validation}, data_{test}\longleftarrow split\_ratio(data_{processed})$
\IF{$\vert S_{searchspace} \vert \equiv 1$}
\STATE $S^* \longleftarrow S \in S_{searchspace}$
\STATE $data_{train}\longleftarrow IS(Concat(data_{train}, data_{validation}), S^*)$
\ELSE

\FOR{$S \in S_{searchspace}$}
\STATE $data_{train} \longleftarrow IS(data_{train}, S)$
\STATE $\pi \longleftarrow TrainModel_{1,000,000}(data_{train}, \mathbfcal{H}^*)$
\STATE $samples.append(Inference(data_{validation}, \pi))$
\ENDFOR
\STATE $S^* \longleftarrow arg\_max(samples.r \in \mathbfcal{R}_{validation})$
\STATE $data_{train}\longleftarrow IS(Concat(data_{train}, data_{validation}, S^*)$
\ENDIF 
\STATE

\FOR{$\{0, 1, ..., test\_sample\_count-1\}$}
\STATE $\pi_{trained} \longleftarrow TrainModel_{1,000,000}(data_{train}, \mathbfcal{H}^*, AdamOptimizer)$
\STATE $\boldsymbol{\pi}_{trained}.append(\pi_{trained})$
\ENDFOR
\STATE \textbf{return} $\boldsymbol{\pi}_{trained}$
\end{algorithmic}
\end{algorithm*}

\begin{algorithm*}[!htb]
\caption{IS($\cdot$, $\cdot$)}
\label{alg:algorithm}
\textbf{Input}: $data_{train}$, $S$ \\
\textbf{Output}: $data_{train}$
\begin{algorithmic}[1] 

\STATE $partition_{size} \longleftarrow floor(\vert data_{train} \vert / S)$
\STATE $i \longleftarrow 0$
\FOR{$j \in \{S-1, S-2,..., 0\}$}
\STATE $data_{train}[i(partition_{size})$:$(i+1)(partition_{size})]$ $\longleftarrow$ $EMA(data_{train}[i(partition_{size}))$:$(i+1)(partition_{size}))],$ $w_l:= j)$
\STATE $i++$
\ENDFOR

\STATE \textbf{return} $data_{train}$
\end{algorithmic}
\end{algorithm*}

\subsection{8. Inference}

The pseudo-code for inference is available in Algorithm 9. Results of PPO-based methods are presented in the main body, Figures 4 and 5. Results of TRPO and A2C-based methods are presented in Appendix Figures 8, 9, 10, and 11.

\begin{figure}[!htb]
    \centering
    \includegraphics[width=\columnwidth]{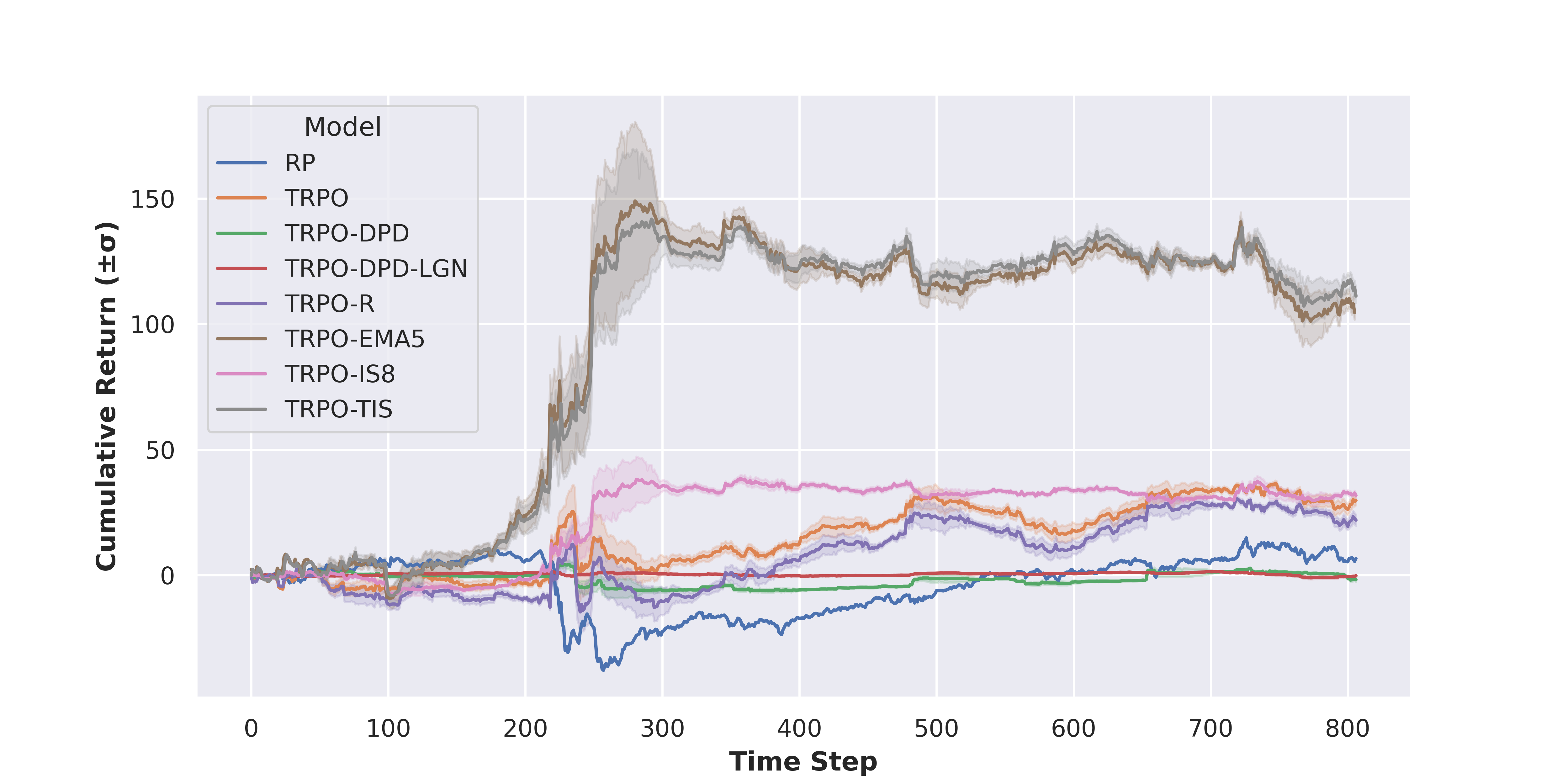}
    \caption{TRPO Test Set Inference (Data Set 1)}
    \label{fig:Test Set Performance}
\end{figure}

\begin{figure}[!htb]
    \centering
    \includegraphics[width=\columnwidth]{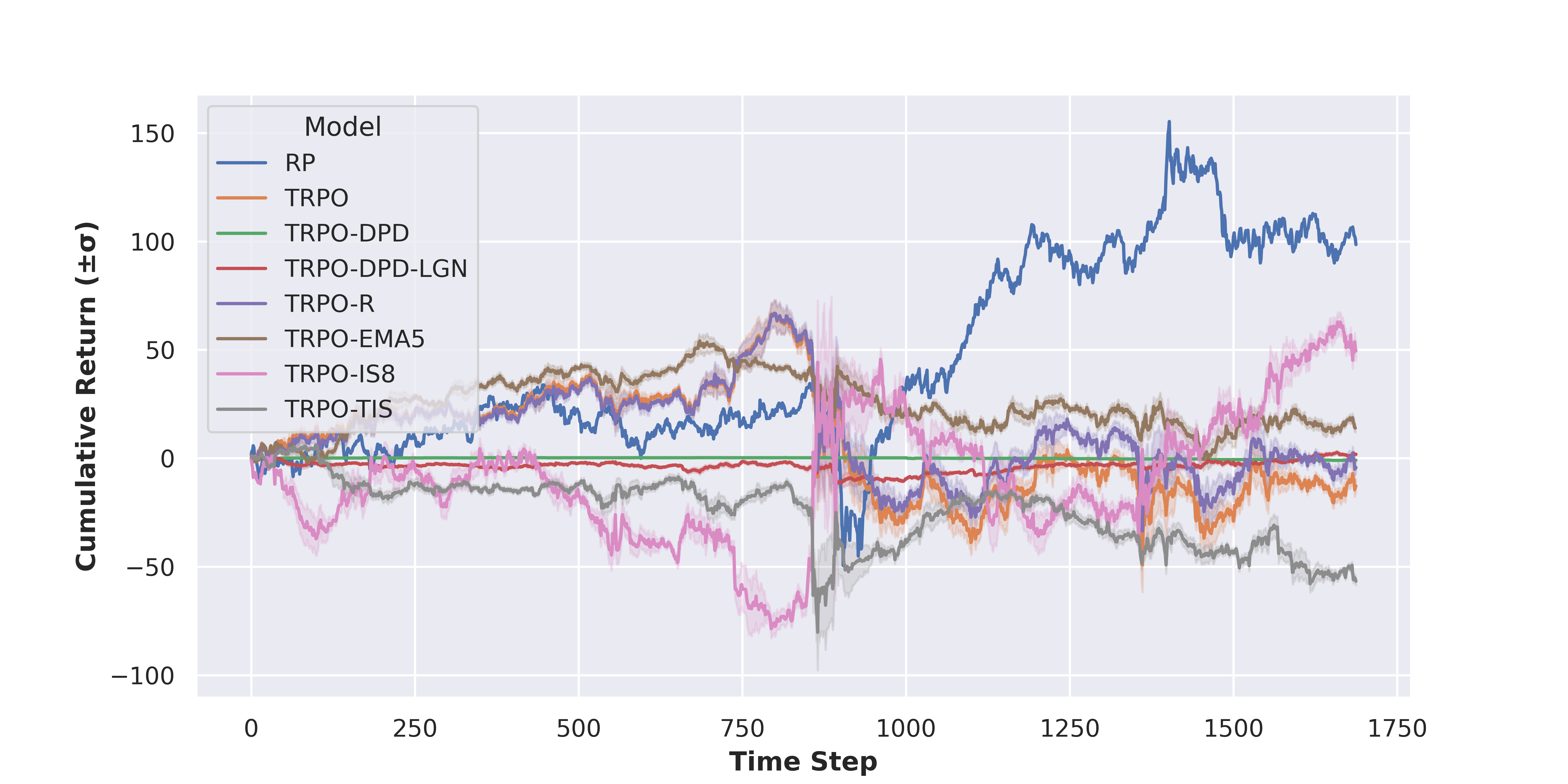}
    \caption{TRPO Test Set Inference (Data Set 2)}
    \label{fig:Test Set Performance}
\end{figure}

\begin{figure}[!htb]
    \centering
    \includegraphics[width=\columnwidth]{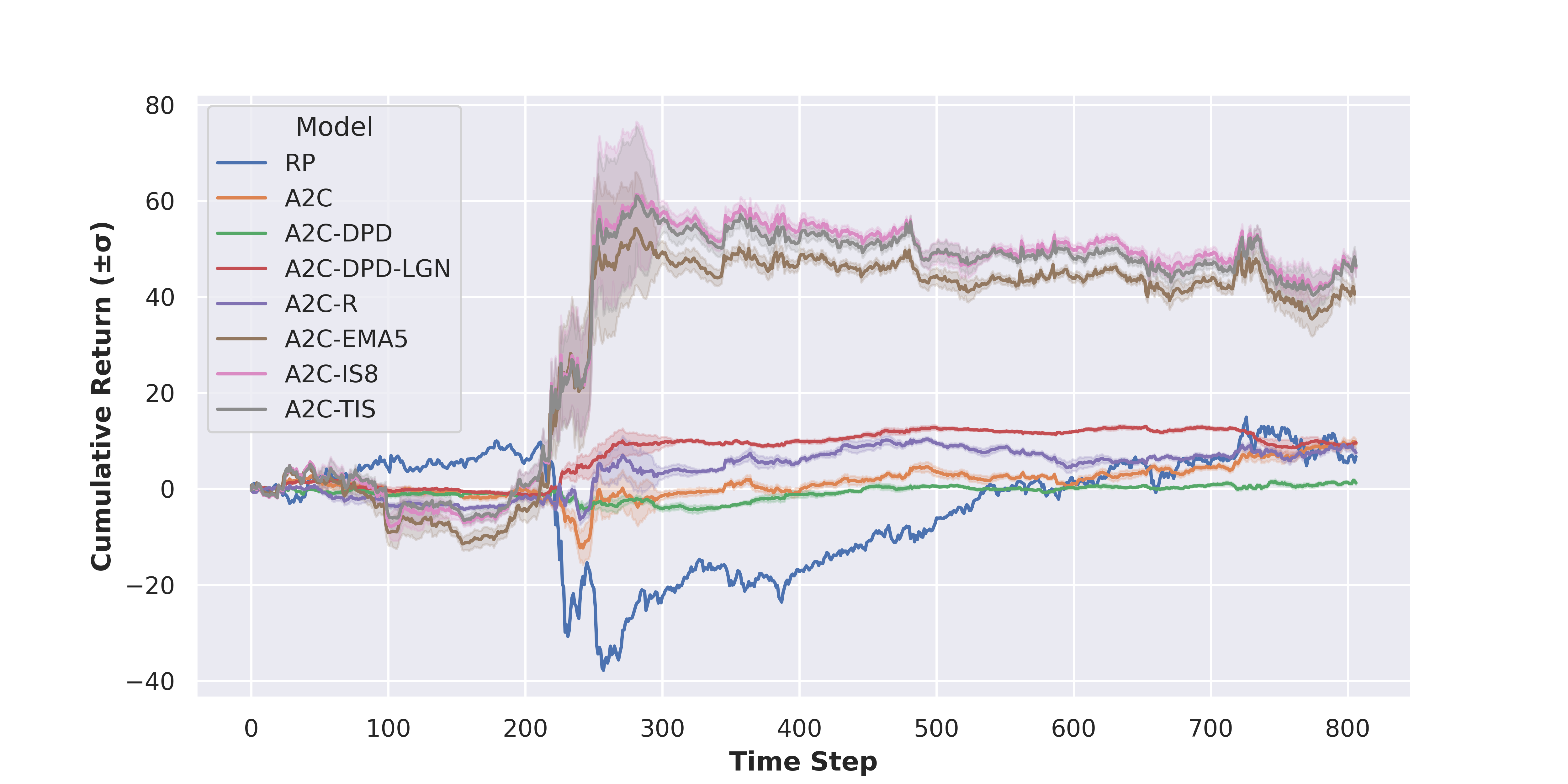}
    \caption{A2C Test Set Inference (Data Set 1)}
    \label{fig:Test Set Performance}
\end{figure}

\begin{figure}[!htb]
    \centering
    \includegraphics[width=\columnwidth]{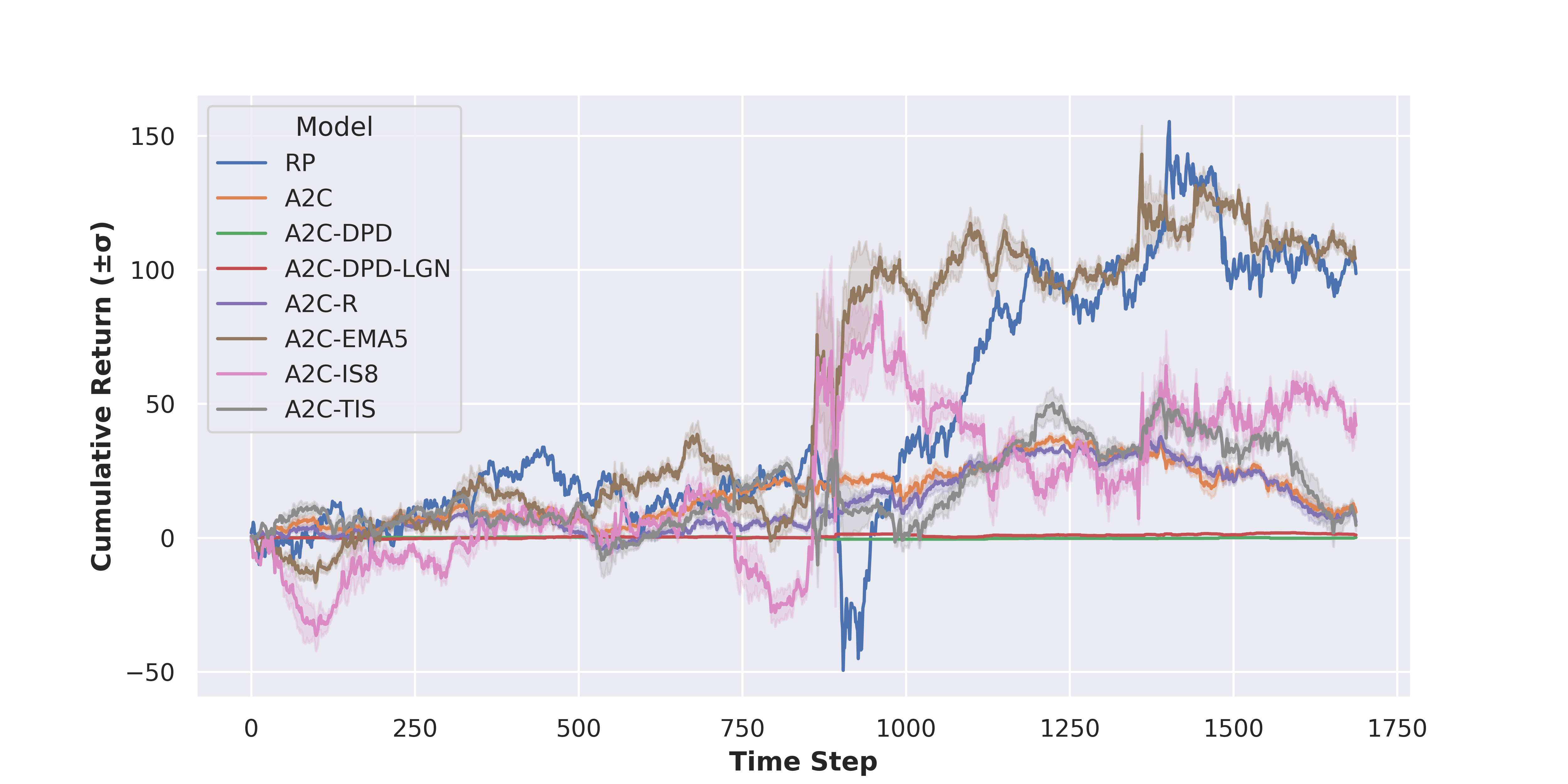}
    \caption{A2C Test Set Inference (Data Set 2)}
    \label{fig:Test Set Performance}
\end{figure}

\begin{algorithm*}[!htb]
\caption{Inference}
\label{alg:algorithm}
\textbf{Input}: $\boldsymbol{\pi}_{trained}$, $MDP$ \\
\textbf{Output}:
\begin{algorithmic}[1] 

\FOR{$\pi \in \boldsymbol{\pi}_{trained}$}

\STATE $\mathbfcal{S} \longleftarrow MDP.reset()$
\STATE
\WHILE{$true$}

\STATE $action, \_states \longleftarrow \pi(\mathbfcal{S})$
\STATE $\mathbfcal{S}, reward, done, \_info \longleftarrow MDP.step(action)$
\IF{$done$}
\STATE $break$
\ENDIF

\ENDWHILE
\STATE

\ENDFOR

\STATE \textbf{return}
\end{algorithmic}
\end{algorithm*}

\subsection{9. Statistical Significance}
Tables 11, 12, and 13 show the one-sided difference T-statistic and corresponding P-values.

\begin{table*}
\centering
  \caption{Statistical Significance against TRPO in Test Set Cumulative Return ($\pm1\sigma$)}
  \label{tab:commands}
  \begin{tabular}{|c|ccc|ccc}
        \toprule
    Method & Data Set 1 & T-stat 1 & P-val 1 (\%) & Data Set 2 & T-stat 2 & P-val 2 (\%)  \\
            \toprule
 \multicolumn{7}{c}{\textbf{Heuristic}} \\
\toprule
    \rowcolor{Gray}
    RP & 2.951 & --- &---& 91.057& --- &--- \\ 
            \toprule
 \multicolumn{7}{c}{\textbf{Baseline}} \\
\toprule
    TRPO & 29.440 $\pm$ 51.712 &---&---& -$12.820 \pm 38.931$ &---&--- \\
\toprule
     \multicolumn{7}{c}{\textbf{Imitation Learning}*} \\
\toprule
   \rowcolor{Gray}
    TRPO-DPD & -1.762 $\pm$ 19.492 &3.99&99.99& -0.9014 $\pm$ 3.239 &-2.16&1.79 \\
    TRPO-DPD-LGN$^\dagger$ & -0.187 $\pm$ 17.948 &3.83&99.98& 1.849 $\pm$ 72.434 &-1.26&10.55\\
\toprule
  \multicolumn{7}{c}{\textbf{Curriculum Learning}*} \\
\toprule
    \rowcolor{Gray}
    TRPO-R$^\ddagger$ & 21.949 $\pm$ 61.999 &0.66&74.33& -4.260 $\pm$ 49.437 &-0.96&16.93\\
    TRPO-EMA5 & 104.599 $\pm$ 44.225 &-7.81&0.00& 13.941 $\pm$ 67.672 &	-2.42&0.88\\   
    \rowcolor{Gray}
    TRPO-IS8& 31.806 $\pm$ 29.014 &-0.28&38.93& \textbf{49.471} $\pm$ \textbf{112.507} &-3.70&0.02\\   
    TRPO-TIS$^\dagger$ & \textbf{111.169} $\pm$ \textbf{44.182} &-8.50&0.00& -56.721 $\pm$ 130.088 &2.29&98.70\\
\bottomrule
\multicolumn{7}{c}{\small *ours, $^\dagger$ablation study, $^\ddagger$na\"{\i}ve approach} \\
  \end{tabular}
\end{table*}

\begin{table*}
\centering
  \caption{Statistical Significance against PPO in Test Set Cumulative Return ($\pm1\sigma$)}
  \label{tab:commands}
  \begin{tabular}{|c|ccc|ccc}
        \toprule
    Method & Data Set 1 & T-stat 1 & P-val 1 (\%) & Data Set 2 & T-stat 2 & P-val 2 (\%)  \\
            \toprule
 \multicolumn{7}{c}{\textbf{Heuristic}} \\
\toprule
    \rowcolor{Gray}    RP & 2.951 & --- &---& 91.057& --- &--- \\
            \toprule
 \multicolumn{7}{c}{\textbf{Baselines}} \\
\toprule
    PPO &29.440 $\pm$ 51.712 &---&---&4.202 $\pm$ 64.185&---&--- \\
   \rowcolor{Gray}
      PPO-OPD & 23.009 $\pm$ 48.459 &0.64 &73.87&-44.461 $\pm$ 177.441 &1.82&96.35 \\
\toprule
     \multicolumn{7}{c}{\textbf{Imitation Learning}*} \\
\toprule
    PPO-DPD & 3.543 $\pm$ 29.152 &3.08 &99.86& -24.805 $\pm$ 79.796& 2.00 &97.60 \\
    \rowcolor{Gray}  PPO-DPD-LGN$^\dagger$ & -3.059 $\pm$ 35.961 & 3.65&99.98 & -4.687 $\pm$ 100.238& 0.53&70.06\\
\toprule
  \multicolumn{7}{c}{\textbf{Curriculum Learning}*} \\
\toprule

    PPO-R$^\ddagger$ &17.028 $\pm$ 50.973  & 1.20& 88.52& -13.728 $\pm$ 69.622&1.33 &90.81\\
     \rowcolor{Gray} PPO-EMA5 & 81.817 $\pm$ 33.423&-6.01&0.00& 35.215 $\pm$ 55.376 & 0.26& 0.56\\   
 PPO-IS8& \textbf{81.873} $\pm$ \textbf{39.850} &-5.68&0.00& \textbf{39.389} $\pm$ \textbf{134.199} &-1.673&4.88 \\   
     \rowcolor{Gray} PPO-TIS$^\dagger$ & 28.032 $\pm$ 53.716 &0.13&55.30& 24.380 $\pm$ 120.121&-1.05&14.91\\
\bottomrule
\multicolumn{7}{c}{\small *ours, $^\dagger$ablation study, $^\ddagger$na\"{\i}ve approach} \\
  \end{tabular}
\end{table*}

\begin{table*}
\centering
  \caption{Statistical Significance against A2C in Test Set Cumulative Return ($\pm1\sigma$)}
  \label{tab:commands}
  \begin{tabular}{|c|ccc|ccc}
        \toprule
    Method & Data Set 1 & T-stat 1 & P-val 1 (\%) & Data Set 2 & T-stat 2 & P-val 2 (\%)  \\
            \toprule
 \multicolumn{7}{c}{\textbf{Heuristic}} \\
\toprule
    \rowcolor{Gray}
    RP & 2.951 & --- &---& 91.057& --- &--- \\
            \toprule
 \multicolumn{7}{c}{\textbf{Baseline}} \\
\toprule
    A2C &9.374 $\pm$ 33.604  &---&---& 9.567 $\pm$ 66.664&---&--- \\
\toprule
     \multicolumn{7}{c}{\textbf{Imitation Learning}*} \\
\toprule
   \rowcolor{Gray}
    A2C-DPD & 1.205 $\pm$ 40.249 & 1.10& 86.33& 0.099 $\pm$ 4.680 &1.00 &83.93 \\
    A2C-DPD-LGN$^\dagger$ &9.613 $\pm$ 41.123  & -0.03 & 48.73& 1.043 $\pm$ 15.055 & 0.88&80.91\\
\toprule
  \multicolumn{7}{c}{\textbf{Curriculum Learning}*} \\
\toprule
    \rowcolor{Gray}
    A2C-R$^\ddagger$ & 7.493 $\pm$ 32.060  & 0.29& 61.24& 5.893 $\pm$ 72.870 & 0.26&60.35\\
    A2C-EMA5 &40.550 $\pm$ 24.504 &-5.30&0.00& \textbf{104.17} $\pm$ \textbf{75.074} & -6.66 & 0.00 \\   
    \rowcolor{Gray}
    A2C-IS8& 46.018 $\pm$ 28.972 &-5.85&0.00& 41.937 $\pm$ 106.385 & -1.82 & 3.60\\   
    A2C-TIS$^\dagger$& \textbf{46.414} $\pm$ \textbf{28.788} &-5.92&0.00& 4.651 $\pm$ 130.721 & 0.24 & 59.34\\

\bottomrule
\multicolumn{7}{c}{\small *ours, $^\dagger$ablation study, $^\ddagger$na\"{\i}ve approach} \\
  \end{tabular}
\end{table*}

\subsection{10. Noise Tolerance Mechanism under Constraints}

We now discuss how our DPD framework, a simple yet pure form of IL, reduces the effects of $\Delta noise$ in the label space when the MDP environment is constrained. These constraints are innately required in financial control as capital is finite, and investment managers are given risk mandates. Our two representative constraints applied to Data Set 1 and Data Set 2, respectively, are available in Equation (4).

First, examine the state of the vanilla (single-learner)---i.e., the input space. $ \forall t \  \exists \mathbfcal{S}_{T'} := \{ \mathbfcal{SEC}^\mathbfcal{S}_{t-1} , ...,  \mathbfcal{SEC}^\mathbfcal{S}_{t-T'} \}$, where $T' \in \mathbb{N}$, $\mathbfcal{SEC}^\mathbfcal{S}_t := \{ \Delta signal^1_t + \Delta noise^1_t , ..., \Delta signal^m_t + \Delta noise^m_t\}$, $m := N$. Similarly, the space that affects the label space, that in turn affects $\pi$ can be described as $ \forall t\ \exists \mathbfcal{L}_T := \{ \mathbfcal{SEC}^\mathbfcal{L}_t, ..., \mathbfcal{SEC}^\mathbfcal{L}_{t+T} \}$, where $T \in \mathbb{N}$, $ \mathbfcal{SEC}^\mathbfcal{L}_t := \{ \Delta signal^1_t + \Delta noise^1_t, ..., \Delta signal^p_t + \Delta noise^p_t\}$, $p \in \mathbb{N}$. $m:= N$ does not necessarily have to equal $p$. $m = N = p$ is a unique case where $\mathbfcal{S} \equiv \mathbfcal{U}$, as in our Commodity Futures MDP environment corresponding to Data Set 2.

The noise tolerance mechanism of $\pi^\psi: \mathbfcal{S}_{T'} \times \mathbfcal{A}^\phi \longmapsto a$ process is best introduced via an example. Suppose  $\mathbfcal{U}:= \{sec_1, ..., sec_4\}$, which $\pi$ expects will change by 1.23, 1.79, 3.01, and -4.02 percent on average, respectively. Now suppose a $\pi$ is asked to give an $a$ vector of size four of discrete values $\in [-1, 0, 1]$ representing whether to take a short, none, or long position $\forall sec$. As the reward (label) is calculated via portfolio return, the $a$ vector is again transformed, $w(a)$, to conform to the constraints. Suppose the constraint is that the position magnitude must always add up to 1---the same constraint given to the first MDP environment, Data Set 1. In our case, we use a simple normalization method to respect the hard constraints. However, another common approach would be to use softmax. Nevertheless, since both approaches are deterministic, the logic holds. In this environment, the deterministic $\mathbfcal{A}^* = [0, 0, 0, -1]$ as allocating the entire portfolio to a short position in $sec_4$ yields the maximum expected return of 4.02 percent.

Now, given that $\pi^\psi$  learns to assign -1 to $sec_4$ and zero to the remaining $sec$, via imitating the $\mathcal{\phi}$, $\exists$ a degree of noise-tolerance, $\mathcal{DNT}$. $\mathcal{DNT}$ is

\[\mathop{rank_{descending}^{i:=0}}_{sec\in\mathbfcal{U}} \vert E[\Delta sec_t] \vert - \mathop{rank_{descending}^{i:=1}}_{sec\in\mathbfcal{U}} \vert E[\Delta sec_t] \vert \tag{6}\], which evaluates to $4.02 - 3.01 = 1.01 $ in our example theoretical case. Translating to an additional $\partial \Delta noise < 1.01$ tolerable $\forall \Delta sec$ without affecting the training of the $\psi$ learner. The formalization of the idea is described below.

\begin{definition}
  Degree of Noise-tolerance, $ \mathcal{DNT} \in \mathbb{R} \ge 0$ is the maximum additional noise tolerable for each element in an expected movement vector s.t. the constraint-respecting weight vector remains unchanged. Formally, it can be described as the following. Given a state-action mapping 
  
\[ \pi_t: \mathbfcal{S}_{T'} \longmapsto a_t \tag{7}\]
\[ s.t.\  \mathop{arg \ min}_{ \pi } -\mathbb{E}( r(\mathbfcal{W}) \vert \mathbfcal{S}, \pi_\theta ) , \tag{8}\]
\newline
and a consequent deterministic transformation of $a$ to $w$, described by $CN$,  where

\[ CN: a_t \longmapsto w_t ,\tag{9}\]
\newline
$s.t. \ w$ respects the constraints of the portfolio environment, $\exists$ by the temporal nature of $\pi_t$, abstracted implicit reasoning of

\[  \mathbb{E}[\Delta sec_t \vert \mathbfcal{S}_{T'} ], \ \forall sec_t \in \mathbfcal{U}, \tag{10}\]
\newline
within the neural-network. Here, $\mathcal{DNT}$ is defined as, $\forall sec \in \mathbfcal{U}$, the following inequality holds:

\[ \partial \vert \mathbb{E}[\Delta sec_t \vert \mathbfcal{S}_{T'} ] \vert < \mathcal{DNT}, s.t. \ \Delta a_t = 0 \tag{11}\]
\newline

where $\partial \vert \mathbb{E}[\cdot] \vert$ expresses any additional noise, and $\Delta a = 0$ expresses no change in action vector, $a$. Naturally, $\Delta w = 0$. 
\end{definition}

This definition is specific for the environment constraints given in this paper. $\mathcal{DNT}$ would be defined differently for different environment constraints. Unlike the vanilla learner, where the $\mathbfcal{R}$ is derived from $\mathbfcal{L}_T$, the student learner's $\mathbfcal{R}$ is derived from the teacher's optimal $\mathbfcal{A}^*$ vector, which has $\mathcal{DNT}$ properties. This does not entirely remove the noise's impact on learning, but does reduce its effect---i.e., the learning system is tolerant to some level of noise. With the example in-hand, a one percent increase of noise in the expected movement of any single $sec$ would change the $r^{vanilla}_t$, but no change in $r^{\psi}_t$ will be observed. In turn, it is synthetically achieving significantly lower $\Delta noise$ in the label space. However, as seen in the main text, this also results in significant removal $\Delta signal$, which is detrimental to learning the parameters.

\end{document}